\documentclass[lettersize,journal]{IEEEtran}
\usepackage{amsmath,amsfonts}
\usepackage{algorithmic}
\usepackage{algorithm}
\usepackage{array}
\usepackage[caption=false,font=normalsize,labelfont=sf,textfont=sf]{subfig}
\usepackage{textcomp}
\usepackage{stfloats}
\usepackage{url}
\usepackage{verbatim}
\usepackage{graphicx}
\usepackage{cite}
\usepackage{caption}
\begin{document}
\title{GAN-based Domain Adaptation for Image-aware Layout Generation in Advertising Poster Design}
\author{Chenchen Xu, Min Zhou, Tiezheng Ge, and Weiwei Xu,~\IEEEmembership{Member,~IEEE}
\thanks{Chenchen Xu is with Anhui Normal University, Wuhu, China; Zhejiang University, Hangzhou, China; and Shenzhen Bay Laboratory, Shenzhen, China (e-mail: xuchenchen@zju.edu.cn).}
\thanks{Min Zhou and Tiezheng Ge are with Alibaba Group, Hangzhou, China (e-mail: yunqi.zm@alibaba-inc.com; tiezheng.gtz@alibaba-inc.com).}
\thanks{Weiwei Xu is with Zhejiang University, Hangzhou, China (e-mail: xww@cad.zju.edu.cn).}
\thanks{Corresponding author: Weiwei Xu.}
}

\maketitle

\begin{abstract}
Layout plays a crucial role in graphic design and poster generation. Recently, the application of deep learning models for layout generation has gained significant attention. This paper focuses on using a GAN-based model conditioned on images to generate advertising poster graphic layouts, requiring a dataset of paired product images and layouts. To address this task, we introduce the Content-aware Graphic Layout Dataset (CGL-Dataset), consisting of 60,548 paired inpainted posters with annotations and 121,000 clean product images. The inpainting artifacts introduce a domain gap between the inpainted posters and clean images. To bridge this gap, we design two GAN-based models. The first model, CGL-GAN, uses Gaussian blur on the inpainted regions to generate layouts. The second model combines unsupervised domain adaptation by introducing a GAN with a pixel-level discriminator (PD), abbreviated as PDA-GAN, to generate image-aware layouts based on the visual texture of input images. The PD is connected to shallow-level feature maps and computes the GAN loss for each input-image pixel. Additionally, we propose three novel content-aware metrics to assess the model's ability to capture the intricate relationships between graphic elements and image content. Quantitative and qualitative evaluations demonstrate that PDA-GAN achieves state-of-the-art performance and generates high-quality image-aware layouts.
\end{abstract}


\section{Introduction}
\IEEEPARstart{G}{raphic} layout is essential for the design of posters, magazines, comics, and webpages. Recently, generative adversarial networks (GANs) have been applied to synthesize graphic layouts by modeling the geometric relationships among different types of 2D elements, such as text and logo bounding boxes~\cite{DBLP:conf/nips/GoodfellowPMXWOCB14,DBLP:conf/iclr/LiYHZX19}. Fine-grained control over the layout generation process can be achieved using conditional GANs. The conditions may include image content and attributes of graphic elements, such as category, area, and aspect ratio~\cite{DBLP:journals/tvcg/LiY0LWX21}. Especially, image content plays an important role in generating image-aware graphic layouts of posters and magazines~\cite{DBLP:journals/tog/ZhengQCL19}.

\label{sec:intro}
    \begin{figure}
    \centering
    \includegraphics[width=8cm]{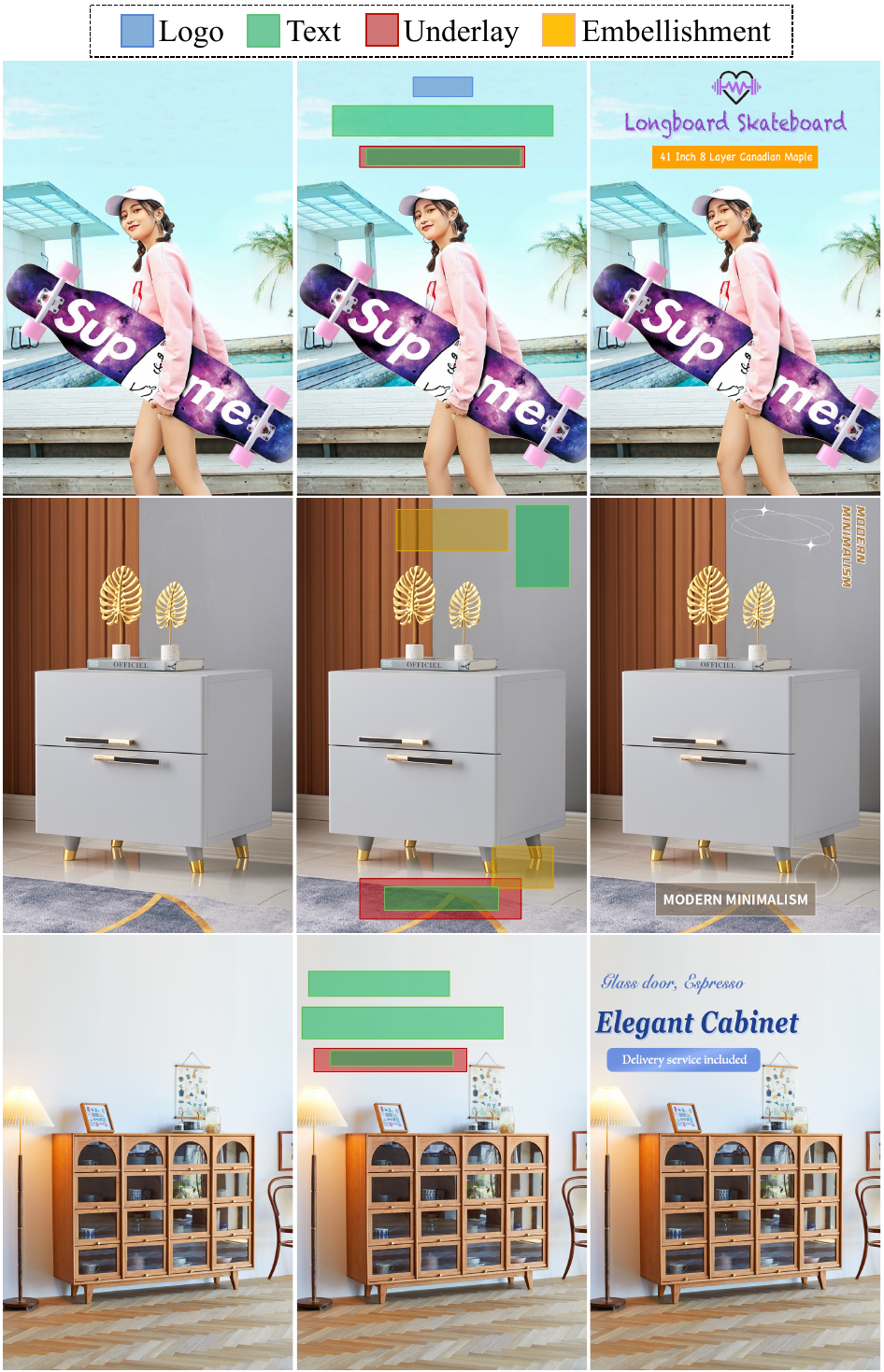}
    \caption{{\bf Examples of image-aware graphic layout generation for advertising posters.} Our model generates graphic layouts (middle) with multiple elements conditioned on product images (left). Designers or even automatic rendering programs can use these layouts to render advertising posters (right).}
    \label{fig:introduction}
    \end{figure}

 This paper focuses on the GAN-based image-aware graphic layout generation for advertising poster design, where the layout involves arranging various elements. As shown in Fig.~\ref{fig:introduction}, our graphic layout generation task involves arranging four classes of elements: logos, text, underlays, and embellishment elements, at appropriate positions based on the product image. The core challenge lies in modeling the relationship between the image content and the layout elements, enabling the neural network to learn how to produce aesthetically pleasing arrangements around the product image. This task can be formulated as a direct set prediction problem, as described in~\cite{DBLP:conf/eccv/CarionMSUKZ20}.

 Constructing a high-quality layout dataset for training image-aware graphic layout generation models is labor-intensive, as it requires professional stylists to design the arrangement of elements to create paired product images and layout data items. To reduce the workload, we collect designed poster images and product images to construct a large content-aware graphic layout dataset (CGL-Dataset), which includes 60,548 advertising posters with annotated layout information and 121,000 clean product images without annotations. The graphic elements imposed on the posters are removed through inpainting~\cite{DBLP:conf/wacv/SuvorovLMRASKGP22} and annotated with their geometric arrangements, resulting in the state-of-the-art CGL-Dataset. While the CGL-Dataset is substantially beneficial to the training of image-aware networks, the inpainting artifact introduces a domain gap between inpainted posters (source domain data) and clean product images (target domain data). 

We propose two approaches to narrow this domain gap. The first approach, CGL-GAN, applies a simple yet effective blurring operation to reduce the distinction between inpainted and non-inpainted regions. This operation smooths both areas simultaneously, narrowing the domain gap and facilitating the generation of image-aware graphic layouts. Although effective, blurred images may lose the delicate color and texture details of products, potentially resulting in undesirable occlusion or placement of graphic elements. The second approach employs an unsupervised domain adaptation technique~\cite{DBLP:conf/iccv/KodirovXFG15,DBLP:conf/iccv/TzengHDS15,DBLP:conf/nips/ZhaoZWMCG18,DBLP:journals/corr/abs-2010-03978,DBLP:journals/pami/JaritzVCWP23} to bridge the domain gap between clean product images and inpainted posters in the CGL-Dataset, significantly improving the quality of generated graphic layouts. The goal of this approach is to align the feature spaces of inpainted posters and clean product images, i.e., the source and target domains. To achieve this, we propose a GAN with a pixel-level domain adaptation discriminator, called PDA-GAN, which allows for fine-grained control over feature space alignment. It is inspired by PatchGAN~\cite{DBLP:conf/cvpr/IsolaZZE17}, but it does not directly adapt to pixel-level alignment in our task.

 The pixel-level discriminator (PD) designed for domain adaptation can replace the Gaussian blurring step, enabling the network to better model the visual and texture details of the product image. The PD is connected to shallow-level feature maps, as the inpainted regions are typically small relative to the whole image and may be difficult to detect at deeper levels with large receptive fields. Additionally, the PD is composed of only three convolutional layers, and its number of parameters is less than $2\%$ of those in the CGL-GAN discriminator. This design significantly reduces the memory and computational cost introduced by the PD. A total of 120,000 target domain images were collected to support the training of PDA-GAN.

 Existing metrics~\cite{DBLP:layoutGAN,DBLP:conf/iccv/JyothiDHSM19,DBLP:conf/cvpr/ArroyoPT21} consider only the relationships among graphic elements and ignore the relationship between graphic elements and image content. Therefore, we propose three novel content-aware metrics to evaluate our methods in terms of image relevance. Considering the particularity and complexity of the image-aware graphic layout generation task, we redesign three content-agnostic graphic metrics. In addition to the conventional Fréchet Inception Distance (FID)~\cite{heusel2017gans,horita2024retrieval}, we further introduce a content-aware variant, termed cFID, as an additional evaluation metric. Experimental results demonstrate that PDA-GAN achieves state-of-the-art (SOTA) performance. It outperforms CGL-GAN (ContentGAN~\cite{DBLP:journals/tog/ZhengQCL19}) on the CGL-Dataset and achieves relative improvements of $6.21\%$ ($26.41\%$), $17.5\%$ ($25.23\%$), $14.5\%$ ($39.81\%$), and $19.07\%$ ($52.89\%$),  on the background complexity, subject occlusion degree, product occlusion degree, and cFID metrics, respectively, leading to significantly improved visual quality of generated graphic layouts in many cases.

This paper is an extension of our previous work presented in~\cite{DBLP:conf/ijcai/ZhouXMGJX22} and~\cite{DBLP:conf/cvpr/XuZGJX23}. Building upon that foundation, we further present a detailed analysis of the CGL-Dataset, a formalized introduction of evaluation metrics, and additional experimental results on domain adaptation. In summary, the main contributions of this work are as follows:
 \begin{itemize}
     \item We contribute a large-scale layout dataset that covers a wide variety of promotional products and professionally designed layouts. To the best of our knowledge, this is the first large-scale dataset focused on advertising poster layout design.
     \item We propose PDA-GAN, which incorporates a novel pixel-level discriminator that operates on shallow-level feature maps to bridge the domain gap between clean product images and annotated inpainted posters in the CGL-Dataset.
     \item To evaluate the global and spatial information of input images learned by the model, we propose three novel content-aware metrics. Given the particularity and complexity of advertising poster graphic design, we redesign three content-agnostic graphic metrics, along with a content-aware version of FID.
 \end{itemize}
 More importantly, both quantitative and qualitative evaluations demonstrate that our method achieves SOTA performance and can generate high-quality image-aware graphic layouts for advertising posters.

\section{Related Work}

\subsection{Image-agnostic Layout Generation}
\noindent Early works~\cite{DBLP:journals/tog/JacobsLSBS03,DBLP:journals/tip/KanungoM03,DBLP:conf/chi/KumarTAK11,DBLP:journals/tog/CaoCL12,DBLP:journals/tvcg/ODonovanAH14,DBLP:journals/tip/HedjamNKC15} often utilize templates and heuristic rules to generate layouts. These approaches rely on professional knowledge and often fail to produce flexible and diverse layouts limited to their hand-crafted rules. LayoutGAN~\cite{DBLP:conf/iclr/LiYHZX19} is the first method to apply generative networks~(in particular, GANs) to synthesize layouts and use self-attention to model element relationships. LayoutVAE~\cite{DBLP:conf/iccv/JyothiDHSM19} and LayoutVTN~\cite{DBLP:conf/cvpr/ArroyoPT21} both follow and apply VAE and autoregressive methods. Diffusion-based models such as LayoutDM~\cite{inoue2023layoutdm} and LayoutDiffusion~\cite{zhang2023layoutdiffusion} have also been proposed for layout generation. Meanwhile, some conditional methods have been proposed to guide the layout generation process~\cite{DBLP:conf/eccv/LeeJELG0Y20,DBLP:conf/cvpr/YangFYW21,DBLP:conf/mm/KikuchiSOY21,DBLP:journals/tvcg/LiY0LWX21,DBLP:conf/iccv/GuptaLA0MS21}. The constraints are in various forms, such as scene graphs, element attributes, and partial layouts. 
However, in general, these methods mainly focus on modeling the internal relationship between graphic elements and rarely consider the relationship between layouts and images.

\subsection{Image-aware Layout Generation}
\noindent In recent years, various image-aware layout generation methods have emerged, including LLM-based~\cite{lin2023layoutprompter} approaches, reflecting the growing interest and rapid progress in this field. In the generation of magazine page layouts, ContentGAN~\cite{DBLP:journals/tog/ZhengQCL19} was the first to model the relationship not only between layout elements but also between layouts and images. However, high-quality training data is relatively scarce, as it requires professional stylists to manually design layouts in order to obtain paired clean images and layout annotations. To address this, ContentGAN uses white patches to mask the graphic elements on magazine pages and employs these processed pages as substitutes for clean images during training. In the context of poster layout generation, we tackle the same issue by applying image inpainting~\cite{DBLP:conf/wacv/SuvorovLMRASKGP22, DBLP:conf/cvpr/HsuHPKZ23} to remove graphic elements from posters, followed by a Gaussian blur applied to the entire image to mitigate inpainting artifacts. While this blur strategy effectively narrows the domain gap between inpainted and clean images, it may also degrade fine color and texture details, leading to suboptimal occlusion or element placement. In this paper, we demonstrate that a pixel-level discriminator designed for domain adaptation can achieve similar goals while avoiding the negative side effects of blurring.

\subsection{Unsupervised Domain Adaptation}
\noindent Unsupervised domain adaptation~\cite{DBLP:journals/corr/abs-2010-03978} aims at aligning the disparity between domains such that a model trained on the source domain
with labels can be generalized into the target domain, which lacks labels. Many related methods~\cite{DBLP:conf/esann/MajumdarN18,DBLP:journals/corr/abs-2205-12923,DBLP:journals/mta/ZhangD21,DBLP:conf/cvpr/Zhang0TL22,DBLP:conf/cvpr/BousmalisSDEK17,DBLP:journals/corr/abs-2010-03978,DBLP:conf/aaai/PeiCLW18,DBLP:journals/pami/GaoZX21,DBLP:journals/tip/RenLZH22,DBLP:journals/pami/JaritzVCWP23,DBLP:journals/pami/NingLXCWZTYY23} have been applied for object recognition, detection and segmentation. Among these methods, \cite{DBLP:conf/cvpr/BousmalisSDEK17,DBLP:journals/corr/abs-2010-03978,DBLP:conf/aaai/PeiCLW18,DBLP:journals/tip/RenLZH22} leverage adversarial domain adaptation approach~\cite{DBLP:series/acvpr/GaninUAGLLML17}. A domain discriminator is employed and outputs a probability value indicating the domain of the input data. In this way, the generator can extract domain-invariant features and bridge the semantic or stylistic gap between the two domains. However, it does not work well when applied directly to our problem, since the inpainted area is small compared to the whole image and is difficult to discern at deep levels. Therefore, we design a pixel-level discriminator, which is connected to shallow-level feature maps and computes GAN loss for each input-image pixel, to effectively solve this.

 \begin{table*}[t!]
     \caption{{\bf{Different dataset of poster layout.}} \textnormal{The symbol $\mathcal{S}$ ($\mathcal{T}$) represents the source (target) domain data.}}
    \label{tab:dataset}
    \centering
    \hspace{-0.2cm}\setlength{\tabcolsep}{1.8mm}{
    \scalebox{1.0}{
    \begin{tabular}{c|c|c|c|c|c|c|c|c}
    \hline
         Dataset    &Status &Train $\mathcal{S}$ &Test $\mathcal{S}$ &Train $\mathcal{T}$  &Test $\mathcal{T}$ &Element classes &Content &Product category \\
    \hline
         NDN~\cite{DBLP:conf/eccv/LeeJELG0Y20}      &Private &0 &500  &0 &0 &Text, logo, image, button &Empty &Car\\
         ICVT~\cite{DBLP:conf/mm/CaoMZLXGJ22} &Private &105,862 &11,762 &0 &166 &Text, logo, underlay &Product &Clothing, electronics, cosmetics, etc.\\
         PKU~\cite{DBLP:conf/cvpr/HsuHPKZ23} &Public &9,974 &0 &0 &905 &Text, logo, underlay &Product &Clothing, electronics, cosmetics, etc.\\
         CGL(Ours)   &Public &54,546 &6,002 &120,000 &1,000 &Text, logo, underlay, embellishment &Product &Clothing, electronics, cosmetics, etc.\\
    \hline
    \end{tabular}
    }
    }

\end{table*}

\begin{figure*}
    \centering
    \hspace{-0.2cm}\includegraphics[width=\textwidth]{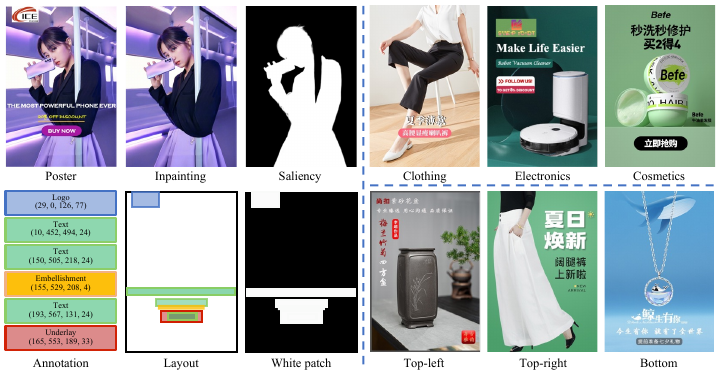}
    \caption{{\bf Examples of CGL-Dataset.} The left part represents the six components of information contained in each sample of the dataset. In the upper right corner are examples of different product types, and in the lower right corner are examples with different layout positions.}
    \label{fig:dataset}
\vspace{-0.0cm}
\end{figure*}

 \begin{figure}[ht]
    \centering
    \hspace{-0.2cm}\includegraphics[width=8.8cm]{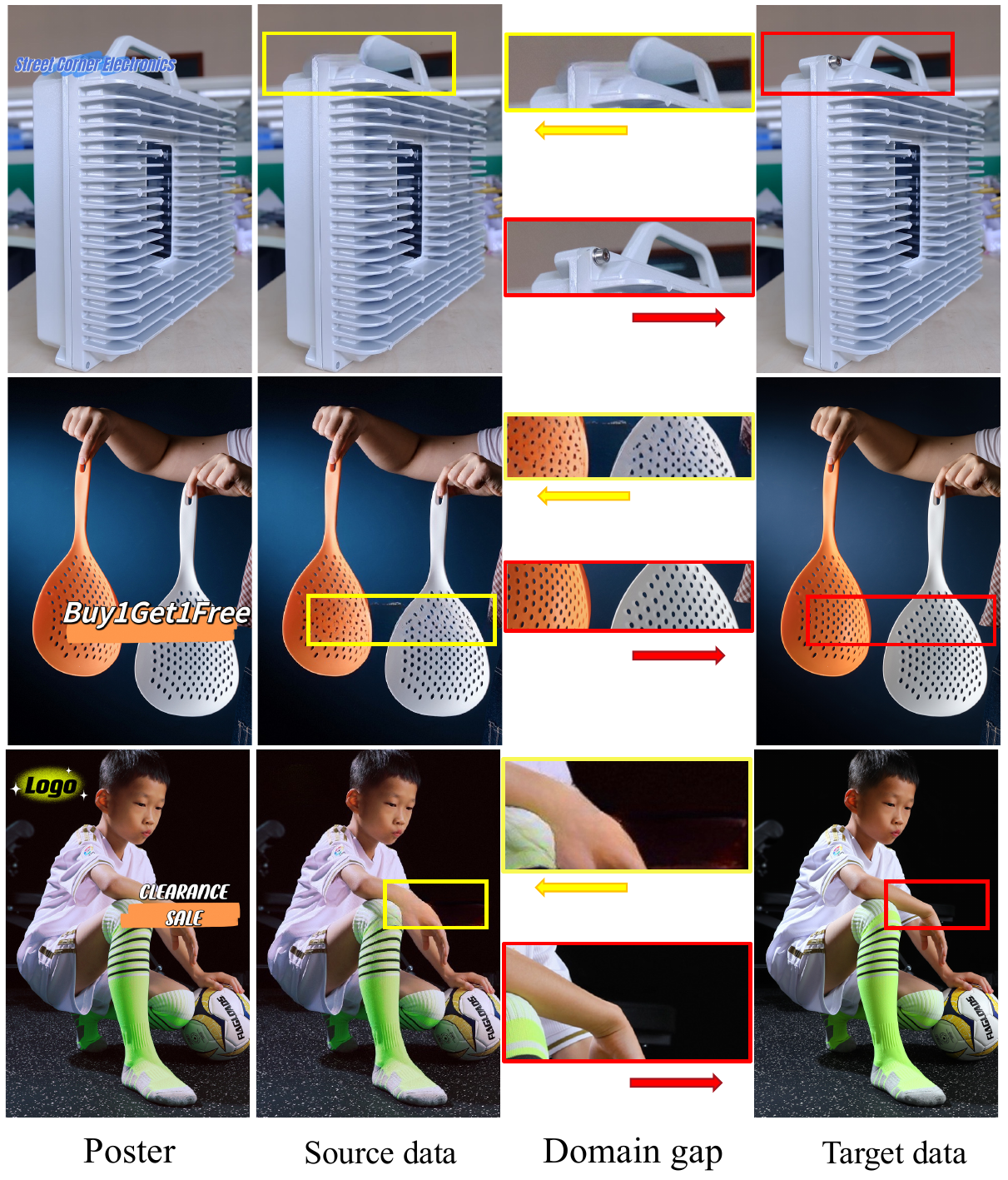}
    \caption{{\bf Domain gap visualization.} To illustrate the domain gap, we manually design several posters based on clean product images (target data), and then inpaint the graphic elements to generate the corresponding source data. The visual content in the inpainted areas (highlighted with yellow boxes) appears distorted and blurred compared to the original content (highlighted with red boxes).}
    \label{fig:Domain gap}
\vspace{-0.0cm}
\end{figure}

\begin{figure*}[t]
\centering
\hspace{0cm}\includegraphics[width=\textwidth]{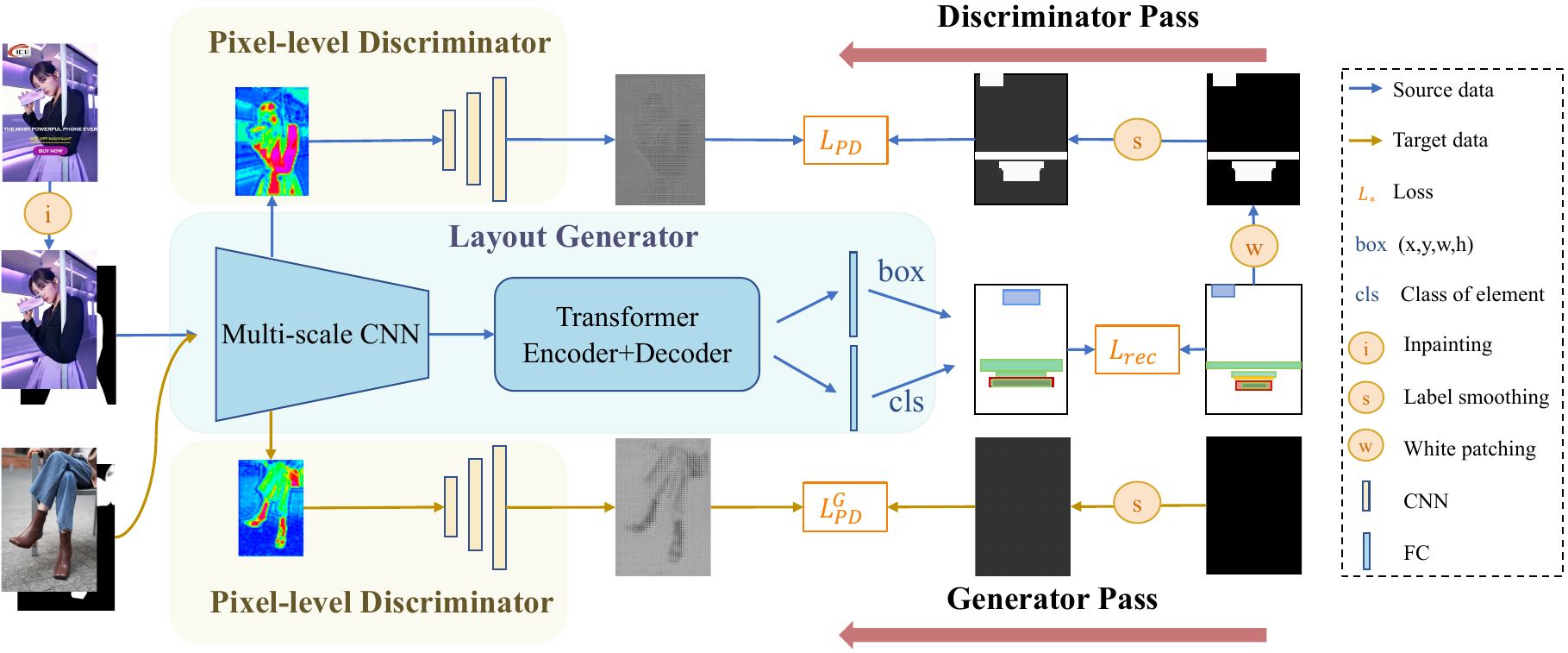}
\caption{{\bf The architecture of our network.} Annotated posters (source domain data) must be inpainted before input to the model. The model has both reconstruction and GAN loss when training with source domain data, while only a GAN loss is used when training with target domain data. Please refer to Sec.~\ref{section.model} for the definition of each loss term: $L_{PD}$, $L_{PD}^G$, and $L_{rec}$. During the discriminator or generator pass, both inpainted and clean images are fed into the discriminator.}
\label{fig:model}
\vspace{-0.0cm}
\end{figure*}

\section{Dataset and Representation}~\label{Datasets}
\vspace{-0.8cm}
\subsection{Different Dataset of Poster Layout}
\noindent Existing publicly available datasets~\cite{DBLP:conf/eccv/LinMBHPRDZ14,DBLP:conf/uist/DekaHFHALNK17,DBLP:conf/uist/LiuCSYMK18,DBLP:conf/icdar/ZhongTJ19} predominantly focus on the relationships between graphic elements, without taking into account the content of the background image. These datasets are inadequate for the task of generating poster layouts. In recent years, a limited number of datasets related to poster layout have been presented, as shown in Tab.~\ref{tab:dataset}. Since some of these datasets are still not publicly available to this day, we have made every effort to gather information from their source papers. 
NDN~\cite{DBLP:conf/eccv/LeeJELG0Y20} presented a banner layout dataset consisting of 500 car advertising posters, designed to validate content-agnostic layout generation methods. This dataset lacks a substantial number of training samples and exhibits a limited variety of poster categories. 
ICVT \cite{DBLP:conf/mm/CaoMZLXGJ22} offers a large-scale dataset of 117,624 poster layouts, to some extent, facilitating the development of image-aware approaches. Unfortunately, this dataset is not publicly accessible and includes only 166 target domain images for testing, which is insufficient to effectively validate model performance.
PKU~\cite{DBLP:conf/cvpr/HsuHPKZ23} released a relatively small poster layout dataset, comprising 9,974 annotated source domain images and 905 target domain images. Similar to ICVT, the layout elements are classified into types of text, logo, and underlay, excluding embellishments, resulting in a lack of diversity.

To this end, we provide and release a substantial dataset named CGL-Dataset, which consists of 60,548 (54,546 for training, 6,002 for test) inpainted posters with annotated layout information and 121,000 (120,000 for training, 1,000 for test) clean product images. It shows advantages in multiple classes of elements (e.g., text, logo, underlay, and embellishment), a variety of products (e.g., clothing, electronics, cosmetics, etc., as shown in the upper right of Fig.~\ref{fig:dataset}), and diversity of layout positions (e.g., top-left, top-right, bottom, etc., as shown in the bottom right of Fig.~\ref{fig:dataset}). From the left part of Fig.~\ref{fig:dataset}, each sample in the source domain comprises five components: poster ${\boldsymbol{x}_{pst}}$, inpainted poster ${\boldsymbol{x}^{inp}_{pst}}$, salient map ${\boldsymbol{x}^{sal}_{pst}}$, layout annotations ${\boldsymbol{l}_{GT}}$, and white-patch map ${\boldsymbol{x}^{wp}_{pst}}$. 
The posters were collected with formal authorization from approved e-commerce platforms, ensuring that all data acquisition complies with copyright regulations. Layout annotations were subsequently performed on these posters in a systematic manner by trained personnel, following consistent labeling standards. The annotation information forms the graphic layout, which consists of $n$ variable-length elements, represented as ${\{\boldsymbol{e}_1, \boldsymbol{e}_2, ..., \boldsymbol{e}_n\}}$. Each element $\boldsymbol{e}$ is represented with its type $c$ and bounding box $b = [x, y, w, h]$. $(x, y)$ represents the top-left coordinates, and $w$ ($h$) represents the width (height) of the bounding box. After obtaining the annotations, we utilized the pretrained InpNet~\cite{DBLP:conf/wacv/SuvorovLMRASKGP22} to perform inpainting on the annotated element box areas, resulting in ${\boldsymbol{x}^{inp}_{pst}}$. Following that, we used pretrained SalNet~\cite{DBLP:conf/aaai/WangCZZ0G20} to extract the salient map ${\boldsymbol{x}^{sal}_{pst}}$ from ${\boldsymbol{x}^{inp}_{pst}}$. Finally, we represent the annotation information in a binary image, values of pixels in element boxes region set to 1 and 0 elsewhere, dubbed white-patch map ${\boldsymbol{x}^{wp}_{pst}}$.

It is worth noting that in the task of image-aware layout generation, we are the first to explicitly recognize and address the domain gap. Consequently, our dataset is the only one that provides a large amount of target domain training data, including clean product images ${\boldsymbol{x}_{img}}$ and corresponding saliency maps ${\boldsymbol{x}^{sal}_{img}}$.
Although background images without any graphic elements may offer a more fundamental basis for layout generation, obtaining such data at scale remains a considerable challenge. This is primarily because clean product images typically require manual design and post-processing by professional designers, making large-scale collection highly labor-intensive and costly. In this work, we instead choose to annotate directly on collected posters, aiming to balance the trade-off between data quality and dataset construction feasibility. Nevertheless, exploring layout generation from clean background images remains a compelling direction for future research.

\subsection{Domain Gap Visualization}
\noindent Different marginal distributions across domains are referred to as domain gap \cite{DBLP:journals/corr/abs-2010-03978}. In this work, the paired images and layouts in the existing dataset  \cite{DBLP:conf/ijcai/ZhouXMGJX22} are collected by inpainting \cite{DBLP:conf/wacv/SuvorovLMRASKGP22} and annotating posters, respectively. A domain gap exists between inpainted posters (source domain data) and clean product images (target domain data).

To illustrate the domain gap, Fig.~\ref{fig:Domain gap} shows examples of source and target domain images.
We selected three clean product images from the target domain, added graphic elements to create posters, and then applied inpainting to these posters to generate the source domain data. This process resulted in distorted and blurred inpainted regions, which constitute a pixel-level domain gap.
 
\section{Our Model} \label{section.model}
\noindent This paper focuses on bridging the domain gap while preserving fine-grained details, such as color and texture, to generate image-aware graphic layouts for poster design.
To achieve this, we introduce two GAN-based models, namely CGL-GAN and PDA-GAN, which share a common generator but employ distinct discriminators. CGL-GAN uses Gaussian-blurred inpainted posters as input to reduce the domain gap. Its discriminator adopts a structure similar to that of~\cite{DBLP:conf/eccv/CarionMSUKZ20}, judging whether the input image–layout pairs are real (annotated layout) or fake (generated layout), and outputs a probability score. Although effective in narrowing the domain gap, blurred images may lose the fine color and texture details of products, leading to unpleasant placement or occlusion of graphic elements.

Therefore, the second approach combines unsupervised domain adaptation techniques to design a GAN with a novel pixel-level discriminator (PD), called PDA-GAN, to generate graphic layouts according to image contents. As shown in Fig.~\ref{fig:model}, our network mainly has two sub-networks: the layout generator network that takes the image (${\boldsymbol{x}^{inp}_{pst}}$ or ${\boldsymbol{x}_{img}}$) and its salient map (${\boldsymbol{x}^{sal}_{pst}}$ or ${\boldsymbol{x}^{sal}_{img}}$) as the input to generate graphic layout and the convolutional neural network for pixel-level discriminator. In this section, we will describe the details of the pixel-level discriminator and the layout generator network, respectively.  

\subsection{Pixel-level Discriminator}
\noindent The design of the pixel-level discriminator is based on the observation that the domain gap between inpainted images ${\boldsymbol{x}^{inp}_{pst}}$ and clean product images ${\boldsymbol{x}_{pst}}$ primarily exists at the pixels synthesized during the inpainting process. Therefore, during the discriminator or generator pass in Fig.~\ref{fig:model}, both inpainted and clean images are fed into the discriminator. When updating the discriminator, we encourage the discriminator to detect the inpainted pixels for ${\boldsymbol{x}^{inp}_{pst}}$ in the source domain. In contrast, when updating the generator, we leverage the pixel-level discriminator to encourage the generator to output shallow feature maps that can fool the discriminator, which means that, even for the feature map computed for ${\boldsymbol{x}^{inp}_{pst}}$, the discriminator's ability to detect inpainted pixels should be weakened fast. In this way, when training converges, the feature space of source and target domain images should be aligned. 

Our pixel-level discriminator network consists of three transposed convolutional layers with filter size $3\times 3$ and stride $2$. Its input is the feature map from the first residual block in the multi-scale CNN. These transposed convolutional layers upsample the feature map, and the final output can be resized to exactly match the dimensions of the input image, facilitating the computation of the discriminator’s training loss.

To calculate the loss $L_{PD}$ for each pixel in the input image, we use the white-patch map to identify whether a pixel has been inpainted. In this map, a pixel value is set to 1 if the corresponding pixel in the input image has been processed by inpainting; otherwise, it is set to 0. For clean images in the target domain, all pixel values in the white-patch map are 0.

When updating the discriminator in the GAN training, the pixel-level discriminator takes shallow-level feature maps as input and outputs a map with one channel whose dimension is consistent with the input image. The loss $L_{PD}$ used to train the discriminator is a mean absolute error (MAE) loss or L1 norm between the white-patch map of input images and the output map. It is defined as:
 \begin{equation}
    \begin{aligned}
    {L}_{PD} = 
    \frac{1}{N_{p}} \sum^{N_{p}}_{i=1} (
    {\left\vert {{\mathbf{P}}^{s,w}_i - {\mathbf{P}}^{s,o}_i} \right\vert} * \alpha \\ + {\left\vert {{\mathbf{P}}^{t,w}_i - {\mathbf{P}}^{t,o}_i} \right\vert} * \beta),
    \end{aligned}
\end{equation}
where $N_{p}$ denotes the number of white-patch map pixels, and $\mathbf{P}_i$ indicates the predicted or ground-truth map for the $i_{\text{th}}$ image. The superscripts of $\mathbf{P}_i$ specify the domain and type: $s$ for source or $t$ for target, and $o$ for prediction or $w$ for ground truth. The two coefficients, $\alpha$ and $\beta$, are employed to strike a balance between the white-patch maps of the source and target domains. Since the area of the inpainted pixels in the white-patch map is usually small, we set the value of $\alpha$ to $2$ and $\beta$ to $1$.

We utilize one-side label smoothing \cite{DBLP:conf/cvpr/SzegedyVISW16,DBLP:journals/corr/Goodfellow17} to improve the generalization ability of the trained model. Since the inpainted areas occupy a small proportion of the input image, we only apply label smoothing for pixels not in the inpainted area (i.e., pixels with a value of 0 in the white-patch map), denoted as one-target label smoothing in our experiments. Precisely, we only set $0$ to $0.2$ in the ground truth white-patch map.

\subsection{Layout Generator}
 \noindent The architecture design of the layout generator network follows the principle of DETR~\cite{DBLP:conf/eccv/CarionMSUKZ20}, which has three modules: a multi-scale convolutional neural network (CNN) used to extract image features~\cite{DBLP:conf/cvpr/HeZRS16,DBLP:conf/cvpr/LinDGHHB17}, a transformer encoder-decoder that takes layout element queries as input to model the relationships between layout elements and the product image~\cite{DBLP:conf/nips/VaswaniSPUJGKP17}, and two fully connected layers to predict the element class and its bounding box based on the element features output by the transformer decoder.

 Concatenated $\boldsymbol{x}^{inp}_{pst}$ with $\boldsymbol{x}^{sal}_{pst}$ (or $\boldsymbol{x}_{img}$ with $\boldsymbol{x}^{sal}_{img}$) is fed into the multi-scale CNN, a ResNet50~\cite{DBLP:conf/cvpr/HeZRS16} whose input channels are changed to four and the part after its final convolutional layer is removed. For the reason that image visual-texture content not only means deep-level semantics such as subject locations, but also includes shallow-level features like region complexity, we introduce a multi-scale strategy on the last two convolutional blocks following FPN~\cite{fpn}. There is a slight difference that we do not generate layouts on each scale separately, like detection networks often do, but concatenate the fused and upsampled features as one. We denote $\boldsymbol{F}_j$ the feature maps of the ${j}$-th convolutional block. Then the multi-scale features can be computed as:
 \begin{equation}
\begin{aligned}
\boldsymbol{F}^{'}_j = \mathbf{Conv_{11}}(\boldsymbol{F}_j)\;;\quad\boldsymbol{F}^{up}_j = \mathbf{Upsample}(\boldsymbol{F}^{'}_j)\;;
\\
\boldsymbol{F}^{fused}_j = \mathbf{Concat}(\boldsymbol{F}^{up}_j, \mathbf{Conv_{33}}(\boldsymbol{F}^{up}_j+\boldsymbol{F}^{'}_{j-1}))
\end{aligned}
\end{equation}
where $\mathbf{Conv_{11}}$, $\mathbf{Conv_{33}}$, $\mathbf{Upsample}$ and  $\mathbf{Concat}$ are network operations of convolution, up-sampling, and concatenation respectively. Although higher-resolution features can provide more details, we empirically find fusing features from the last two blocks yields good results and is more efficient for training.

The multi-scale features are projected into ${d}$ channels and then flattened by channel as the input to the transformer encoder. The encoder uses a standard transformer architecture to further refine the image features. The decoder utilizes cross-attention to learn the relationship between the image content and the graphic layouts. Both the encoder and decoder have six layers, and the hidden dims are ${d}$. Position encodings are added in both the encoder and decoder. Finally, the decoder features of each element are passed through two fully connected layers (FCs) to predict the corresponding class ${\boldsymbol{c}}$ and box coordinates ${\boldsymbol{b}}$ logits. The final class and box results are computed using the softmax and sigmoid functions.

 When updating the generator network in the GAN training, we aim to fool the updated discriminator in the detection of inpainted pixels. Therefore, the loss $L_{PD}$ is modified to penalize the generator network if the discriminator outputs a pixel value of $1$. Thus, we have:
  \begin{equation}
    \begin{aligned}
    {L}^{G}_{PD} =
    \frac{1}{N_{p}} \sum^{N_{p}}_{i=1} (
    {\left\vert {\hat{\mathbf{P}}^{s}_i - {\mathbf{P}}^{s,o}_i} \right\vert} * \alpha \\ + {\left\vert {{\mathbf{P}}^{t,w}_i - {\mathbf{P}}^{t,o}_i} \right\vert} * \beta),
    \end{aligned}
\end{equation}
where the values of pixels in $\hat{\mathbf{P}}^{s}_i$ are all set to $0.2$. 
The training loss for the layout generator network is as follows:
\begin{equation} \label{eq3}
    {L}_{G} = L_{rec} + \gamma * L^{G}_{PD},
\end{equation}
where the value of the weight coefficient $\gamma$ is set to 6, and the $L_{rec}$ is the reconstruction loss to penalize the deviation between the graphic layout generated by the network and the annotated ground-truth layout for the inpainted images in the source domain. We calculate the reconstruction loss $L_{rec}$ as the direct set prediction loss in \cite{DBLP:conf/eccv/CarionMSUKZ20}. 

\section{Metrics}
\noindent To better evaluate the performance of poster layout generation, we propose three novel content-aware metrics: $R_{com}$, $R_{shm}$, and $R_{sub}$, which assess background complexity, subject occlusion, and product occlusion, respectively. These content-aware metrics are particularly important in the context of this research. In addition, we introduce cFID, a content-aware variant of the standard FID~\cite{heusel2017gans,horita2024retrieval}, to more effectively evaluate the semantic alignment between layout structures and image content. We also introduce four conventional graphic metrics: $R_{ove}$, $R_{und}$, $R_{ali}$, which quantify layout overlap, underlay overlap, and element alignment, respectively, along with FID~\cite{DBLP:conf/eccv/LeeJELG0Y20,horita2024retrieval}, which evaluates the overall distributional similarity between generated and real layouts. Moreover, we include the metric $R_{occ}$ to measure the ratio of non-empty layouts predicted by the models. In this section, we will provide formal definitions of these metrics and use them to demonstrate the effectiveness of our approach.

\subsection{Content-aware Metrics}
\noindent Since there are no existing metrics specifically designed for the task of image-aware layout generation, we propose three novel content-aware metrics and a redesigned version of FID, termed cFID, tailored for image-aware layout evaluation. The metric $R_{com}$ is designed to evaluate the visual complexity of the background in the regions where text elements (without underlays) are placed, which may affect text readability. A visually cluttered background—such as one with strong edges, textures, or noise—can reduce the clarity of overlaid text and impair the viewer's ability to quickly perceive the text information.
To quantify such background complexity, we compute the average gradient magnitude within the text-only bounding boxes. Specifically, for each pixel, the gradients along the $x$ and $y$ directions are computed using the Sobel operator, and the magnitude (i.e., the Euclidean norm) of the gradient vector is calculated. The final score $R_{com}$ is defined as the average gradient magnitude over all pixels within the predicted text regions. A higher $R_{com}$ indicates more complex background textures and thus potentially lower text readability. The metric $R_{com}$ is computed as: 
\begin{equation}
R_{com} = \frac{1}{N} \sum_{i=1}^{N} \left( \frac{1}{|R_i|} \sum_{p \in R_i} \left\| \nabla \boldsymbol{x}_{p} \right\|_2 \right)
\end{equation}
where $N$ is the number of text-only elements, and $|R_i|$ denotes the number of pixels within the $i$-th text region. $\nabla \boldsymbol{x}_p$ denotes the Sobel gradient vector at pixel $p$, and $\left\| \nabla \boldsymbol{x}_p \right\|_2$ represents its $\ell_2$-norm, i.e., the magnitude of the gradient.

The metric $R_{shm}$ is proposed to evaluate how severely graphic elements occlude the primary subject or product within the background image. A key aesthetic principle in advertising poster design is the clear visibility of core visual content (e.g., people or products), which should remain visually salient after the layout is applied. Therefore, a lower $R_{shm}$ indicates less visual interference from layout elements and better preservation of the original salient regions.
To quantify the semantic impact of layout occlusion, we compute the perceptual difference between the saliency map before and after layout masking. Specifically, we extract high-level semantic representations from a pretrained VGG16~\cite{DBLP:journals/corr/SimonyanZ14a} network by feeding it two saliency-based images: one is the original salient map $x^{sal}$, and the other is the same map with layout elements masked out, denoted as $(x^{sal}, y^l)$. The Euclidean distance between the corresponding output logits from VGG16 reflects how much the layout occludes semantically important content in the image:
\begin{equation}R_{shm} =  
L_2[ \boldsymbol{VGG}(x^{sal}), \boldsymbol{VGG}(x^{sal},y^l) ].
\end{equation}
Here, $x^{sal}$ represents the input saliency map, and $y^l$ denotes the binary mask of the predicted layout. This distance reflects how much the layout disrupts the semantic content captured by the saliency map.

To compute $R_{sub}$, we aim to evaluate how much the layout design interferes with the recognizability of products. Unlike general saliency maps that highlight visually prominent areas, we leverage CLIP-based attention maps conditioned on product category tags, which capture semantic relevance between image regions and textual queries. Specifically, we extract product category names from product pages and use them as text prompts to query attention maps from a pretrained CLIP model\footnote{\url{https://github.com/hila-chefer/Transformer-MM-Explainability}}~\cite{DBLP:conf/icml/RadfordKHRGASAM21,DBLP:conf/iccv/CheferGW21}. The attention values within each layout bounding box are summed to reflect potential occlusions over semantically important regions.
\begin{equation} R^s_{sub} = \frac{1}{N} \sum_{i=1}^{N} \boldsymbol{Rgn}(\boldsymbol{CLIP}({Map}^s)), \end{equation} where $R^s_{sub}$ denotes the $R_{sub}$ value for sample $s$, ${Map}^s$ is the CLIP attention map for $s$, and $\boldsymbol{Rgn}$ extracts attention values within the predicted layout regions. A lower $R_{sub}$ score suggests that layout elements better preserve semantically critical product areas. Compared to general saliency, CLIP attention maps are used here because they better reflect concept-level localization and can handle diverse products that may not be visually salient but are semantically important. 

Fréchet Inception Distance (FID)~\cite{heusel2017gans,horita2024retrieval} is widely used to evaluate the similarity between two image distributions based on deep features. However, since our task focuses on image-aware layout generation and the test set lacks ground-truth layout annotations, the standard FID between predicted and real layouts cannot be directly applied. To address this, we propose a redesigned version of FID, termed cFID, to indirectly assess layout–image harmony. Specifically, we first use an InpNet~\cite{DBLP:conf/wacv/SuvorovLMRASKGP22} to erase the regions occupied by predicted layout elements on test images. Then, we compute the cFID value between the original image set $\boldsymbol{x}_{img}$ and the inpainted image set $\boldsymbol{x}_{img}^{inp}$ to evaluate the semantic disruption caused by the layouts. A lower value of this image-aware layout cFID indicates better preservation of image semantics and better coordination between layout and image content. 
The cFID is computed as follows:
\begin{equation}\label{EQcFID}
    \begin{aligned}
    \mathrm{cFID} = \ 
    & \left\| \mu_{\boldsymbol{x}_{img}} - \mu_{\boldsymbol{x}^{inp}_{img}} \right\|_2^2 \\
    & + \mathrm{Tr} \big( \Sigma_{\boldsymbol{x}_{img}} + \Sigma_{\boldsymbol{x}^{inp}_{img}} - 2 \cdot ( \Sigma_{\boldsymbol{x}_{img}} \Sigma_{\boldsymbol{x}^{inp}_{img}} )^{\frac{1}{2}} \big)
    \end{aligned}
\end{equation}
where $\mu_{\boldsymbol{x}_{img}}$ and $\Sigma_{\boldsymbol{x}_{img}}$ are the mean and covariance of the Inception features for the original images, and $\mu_{\boldsymbol{x}_{img}^{inp}}$ and $\Sigma_{\boldsymbol{x}_{img}^{inp}}$ are the mean and covariance for the inpainted images.

\subsection{Graphic Metrics}
\noindent Content-agnostic graphic metrics, such as overlap and alignment of layout elements, are described in~\cite{DBLP:journals/pami/LiYHZX21,DBLP:journals/tvcg/LiY0LWX21}. However, these metrics overlook the different element types and relationships between different types of elements in the image-aware layout generation task. As mentioned in the paper, the graphic layouts of advertising posters encompass four types of elements: logos, texts, underlays, and embellishments. Specifically, the underlay elements are allowed to overlap with any other elements to improve the readability of texts, since it is possible that the desirable color of text is not salient on a background image. The layout also allows an embellishment to overlap with other elements, except for other embellishments, as embellishments are typically used to decorate posters to enhance the aesthetics.

We follow the equation in \cite{DBLP:journals/tvcg/LiY0LWX21} to calculate the layout overlap metric as follows:
\begin{equation}\label{eq4}R^{\mathbf{e}}_{ove} = \sum_{i\in{\mathbf{e}}}\sum_{\left(j\in{\mathbf{e}}\right)\ne{i}}{\frac{a_{i}\cap{a_{j}}}{a_{i}}}, \end{equation}
$a_i$ means the area of the $i$th box in the set of elements bounding boxes $\mathbf{e}$. In this work,  $\mathbf{e}$ refers to the set of elements bounding boxes of the $\{logo, text\}$ or $\{embellishment\}$.
\begin{equation}R_{ove} = R_{ove}^{\{logo,text\}} + R_{ove}^{\{embellishment\}}\end{equation}

The underlay, as a background element, is primarily used to emphasize or highlight another visual element. Thus, it is expected that an underlay should not appear alone but be overlaid by at least one other type of element (e.g., text or logo). To evaluate the degree to which underlay elements are functionally utilized, we define the underlay overlap degree metric $R_{und}$. This metric quantifies the extent to which underlay regions are overlapped by other content elements. Specifically, for each underlay element, we compute its maximal overlap ratio with elements of other types and then sum the results over all underlays. A higher $R_{und}$ value indicates that underlays are more effectively used to support visible content. Although this formulation might seem to favor cases where underlays are closely fitted to overlaid elements, its actual purpose is to assess whether underlays are being effectively used to support other visual content. If an underlay is largely covered by meaningful elements such as text or logos, it indicates that the layout is making effective use of the underlay to enhance visual emphasis, which leads to a higher score. Conversely, underlays that remain mostly uncovered are considered less functional and result in lower scores. We adapt Eq.~\ref{eq4} to calculate $R_{und}$ as follows:
\begin{equation}R_{und} = \sum_{i\in{\mathbf{e}_1}}\max_{j\in{\mathbf{e}_2}}\frac{a_i\cap{a_j}}{a_i}\end{equation}
$$\mathbf{e}_1 = \{underlay\}$$
$$\mathbf{e}_2 = \{logo, text, embellishment\},$$
where $a_i$ means the area of the $i$th box in $\mathbf{e}_1$, and $\mathbf{e}_1$ and $\mathbf{e}_2$ represent the set of elements.

The metric $R_{ali}$ is used to demonstrate that elements in an aesthetic graphic layout tend to align in one dimension. We follow the equation in \cite{DBLP:journals/tvcg/LiY0LWX21} to calculate the alignment distance of elements:
\begin{equation}
R_{ali} = \sum_{i=1}^N \min{\left(\mathcal{D}{x_i^l}, \mathcal{D}{x_i^c}, \mathcal{D}{x_i^r}, \mathcal{D}{y_i^t}, \mathcal{D}{y_i^c}, \mathcal{D}{y_i^b}\right)}
\end{equation}
$N$ represents the total number of predicted elements. $\mathcal{D}{x_i^l}, \mathcal{D}{x_i^c}, \mathcal{D}{x_i^r}, \mathcal{D}{y_i^t}, \mathcal{D}{y_i^c}, \mathcal{D}{y_i^b}$ represent the minimum distance between the $i$th bounding box and other bounding boxes in the dimensions of the left, horizontal midpoint, right, top, vertical midpoint, and bottom, respectively. $\mathcal{D}{x_i^*\left(* = l, c, r\right)}$ refers to \cite{DBLP:journals/tvcg/LiY0LWX21} as: 
\begin{equation}\mathcal{D}{x_i^* = \min_{{j}\ne{i}}}\left\vert{x_i^* - x_j^*}\right\vert\end{equation}
$\mathcal{D}{y_i^*\left(* = t, c, b\right)}$ can be calculated similarly.

To enable computation of the conventional FID~\cite{heusel2017gans}, which requires ground-truth layout annotations, the test set of 6,002 annotated layout-image pairs from the CGL-Dataset is utilized. For each image, the model generates a corresponding layout. FID is then computed between the distributions of real and generated layout features. 
Similar in form to Eq.~\ref{EQcFID}, FID is calculated as follows:
\begin{equation}
    \begin{aligned}
    \mathrm{FID} = \ 
    & \left\| \mu_{\boldsymbol{l}_{GT}} - \mu_{\boldsymbol{l}_{pre}} \right\|_2^2 \\
    & + \mathrm{Tr} \big( \Sigma_{\boldsymbol{l}_{GT}} + \Sigma_{\boldsymbol{l}_{pre}} - 2 \cdot ( \Sigma_{\boldsymbol{l}_{GT}} \Sigma_{\boldsymbol{l}_{pre}} )^{\frac{1}{2}} \big)
    \end{aligned}
\end{equation}
where $\mu_{\boldsymbol{l}_{GT}}$, $\Sigma_{\boldsymbol{l}_{GT}}$ and $\mu_{\boldsymbol{l}_{pre}}$, $\Sigma_{\boldsymbol{l}_{GT}}$ denote the mean and covariance of layout features extracted from the real and generated layouts, respectively.

 \begin{table*}[t!]
    \caption{{\bf{Comparison with image-aware methods.}} \textnormal{Bold and underlined numbers denote the best and second best respectively. $\downarrow$ (or $\uparrow$) means the smaller (or bigger) value, the better.}}
    \label{tab:content-aware}
    \centering
    \setlength{\tabcolsep}{2.28mm}{
    \scalebox{1.18}{
    \begin{tabular}{lcccc|ccccc}
    \hline
         Model    &$R_{com}\downarrow$ &$R_{shm}\downarrow$ &$R_{sub}\downarrow$ &$cFID\downarrow$ &$R_{ove}\downarrow$ &$R_{und}\uparrow$ &$R_{ali}\downarrow$ &$FID\downarrow$ &$R_{occ}\uparrow$\\
    \hline
         ContentGAN~\cite{DBLP:journals/tog/ZhengQCL19}      &45.59 &17.08 &1.143 &26.30 &0.0397 &0.8626 &\underline{0.0071} &9.62 &93.4\\
         Layoutprompter~\cite{lin2023layoutprompter}                    &39.77 &\underline{14.66} &0.840 &25.55 &0.2251 &0.6786 &\bf{0.0068} &8.67 &99.5\\
         CGL-GAN(Ours) &\underline{35.77} &15.47 &\underline{0.805} &\underline{15.31} &\bf{0.0233} &\underline{0.9359} &0.0098 &\underline{5.10} &\underline{99.6}\\
         PDA-GAN(Ours)   &\bf{33.55} &\bf{12.77} &\bf{0.688} &\bf{12.39} &\underline{0.0290} &\bf{0.9481} &0.0105 &\bf{4.98} &\bf{99.7}\\
    \hline
    \end{tabular}
    }
    }
\end{table*}

\begin{table}
    \caption{{\bf{User study.}} \textnormal{$P_e$ ($P_b$) represents the percentage of eligible-selected (best-selected) layouts. The symbol * denotes the professional group.}}
    \label{tab:user study}
    \centering
    \setlength{\tabcolsep}{2mm}{
    \scalebox{1.2}{
    \begin{tabular}{l|cccc}
    \hline
         Model  &$P_{e}\uparrow$ &$P_{b}\uparrow$  &$P^{*}_{e}\uparrow$ &$P^{*}_{b}\uparrow$\\ 
    \hline
         ContentGAN~\cite{DBLP:journals/tog/ZhengQCL19} &18.89 &19.36 &18.25 &16.58\\
         Layoutprompter~\cite{lin2023layoutprompter} &26.25  &20.12 &24.58 &18.24\\
         CGL-GAN &26.12 &28.36 &26.45 &26.46\\
         PDA-GAN &\bf{28.74} &\bf{32.16} &\bf{30.72} &\bf{38.72}\\
    \hline
    \end{tabular}
    }
    }
\end{table}

 \begin{figure*}
    \centering
    \hspace{0cm}\includegraphics[width=\textwidth]{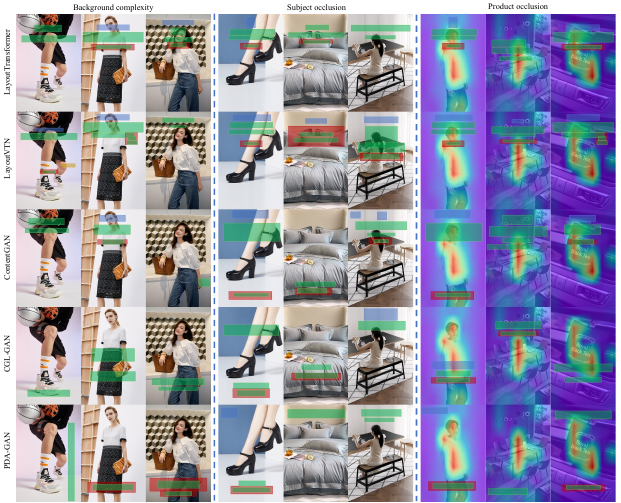}
    \caption{{\bf Qualitative evaluation for different models.} Layouts in each column are conditioned on the same image, while those in each row are generated by the same model. This figure provides a qualitative comparison and analysis of different models from three perspectives: background complexity of text elements, subject overlap, and product overlap attention maps.}
    \label{fig:3}
\end{figure*}

\begin{table}
    \caption{{\bf{Comparison with image-agnostic methods.}} \textnormal{$LT$ and $VTN$ represent LayoutTransformer and LayoutVTN, respectively.}}
    \label{tab:content-unaware}
    \centering
    \setlength{\tabcolsep}{0.5mm}{
    \scalebox{0.938}{
    \begin{tabular}{lcccc|ccccc}
    \hline
         Model   &$R_{com}\downarrow$ &$R_{shm}\downarrow$ &$R_{sub}\downarrow$  &$cFID\downarrow$ &$R_{ove}\downarrow$ &$R_{und}\uparrow$  &$R_{ali}\downarrow$ &$FID\downarrow$\\
    \hline
         LT~\cite{DBLP:conf/iccv/GuptaLA0MS21}     &40.92 &21.08 &1.310 &27.24 &0.0156 &0.9516 &0.0049 &6.25\\
         VTN~\cite{DBLP:conf/cvpr/ArroyoPT21}      &41.77 &22.21 &1.323 &30.14 &0.0130 &\bf{0.9698} &0.0047 &7.13\\
         LDM~\cite{inoue2023layoutdm}      &41.20 &27.91 &1.792 &32.36 &0.0146 &0.9532 &0.0032 &5.02\\
         LD~\cite{zhang2023layoutdiffusion}      &40.56 &27.36 &1.772 &28.76 &\bf{0.0116} &0.9624 &\bf{0.0028} &\bf{4.28}\\
         CGL-GAN &35.77 &15.47 &0.805 &15.31 &0.0233 &0.9359 &0.0098 &5.10 \\
         PDA-GAN     &\bf{33.55} &\bf{12.77} &\bf{0.688} &\bf{12.39} &0.0290 &0.9481 &0.0105 &4.98\\
    \hline
    \end{tabular}
    }
    }
\end{table}

\section{Experiments}
\noindent In this section, we first introduce the implementation details of our experimental setup. Next, we present both quantitative and qualitative comparisons to demonstrate that our model achieves SOTA performance. We then provide visual examples to show how PDA-GAN effectively bridges the domain gap. Finally, we conduct ablation studies to evaluate the contribution of each component to the quality of the generated layouts, and extend our experiments to explore the effect of the training dataset, natural language-guided layout generation, and failure cases from alternative discriminator designs.

\subsection{Implementation Details}
\noindent We implement our model using PyTorch 1.7.1 and train it with the Adam optimizer~\cite{DBLP:journals/corr/KingmaB14}. The initial learning rate is set to $10^{-5}$ for the generator backbone and $10^{-4}$ for the remaining parts of the model. Training is conducted for 300 epochs with a batch size of 128, and all learning rates are reduced by a factor of 10 after 200 epochs. For fair comparison in the ablation studies, inpainted posters and product images are resized to $240 \times 350$ before being used as input.

We observe that, during training, the network is prone to bias towards source domain data. It might be due to the additional reconstruction loss for the source domain to supervise the generator of the model. Therefore, to balance the influence of the two domains, 8000 samples are randomly selected from the CGL-Dataset as the source domain data. In each epoch, the 8000 source domain samples are processed, and another 8000 samples of the target domain images are randomly selected. We refer to this choice of training data as Data I. If all the CGL-Dataset training images are used for comparison, we refer to it as Data II. In the following, if not clearly mentioned, PDA-GAN (CGL-GAN) is trained with Data I (Data II). The total training time for PDA-GAN on Data II is about 8 hours, while CGL-GAN on Data I takes approximately 43.5 hours, both utilizing 16 NVIDIA V100 GPUs.

 \subsection{Comparison with State-of-the-art Methods}\label{Experiments_SOTA}
 \noindent\textbf{Layout generation with image contents.} 
 We begin by conducting experiments to compare PDA-GAN with ContentGAN~\cite{DBLP:journals/tog/ZhengQCL19}, Layoutprompter~\cite{lin2023layoutprompter}, and CGL-GAN, which are capable of generating image-aware layouts. Quantitative results can be seen from Tab.~\ref{tab:content-aware}. Our models, PDA-GAN and CGL-GAN, have demonstrated strong performance across the majority of metrics. Notably, PDA-GAN achieves the best results in most metrics, especially in the content-aware metrics, since PDA-GAN preserves the image color and texture details. For example, PDA-GAN outperforms ContentGAN, Layoutprompter, and CGL-GAN by $26.4\%$, $15.6\%$, and $6.21\%$, respectively, with respect to background complexity $R_{com}$.
 As shown in the first column in Fig.~\ref{fig:3}, compared with those by ContentGAN and CGL-GAN, bounding boxes of text elements generated by PDA-GAN are more likely to appear in simple background areas, which improves the readability of the text information. As shown in the second and third columns, when the background of the text element is complex, PDA-GAN will generate an underlay bounding box to replace the complex background to enhance the readability of text information.

 Compared to ContentGAN, Layoutprompter, and CGL-GAN, PDA-GAN reduces the subject occlusion degree $R_{shm}$ by $25.2\%$, $12.9\%$, and $17.5\%$, respectively. From the three middle columns of Fig.~\ref{fig:3}, for ContentGAN or CGL-GAN, the presentation of the subject content information is largely affected since the generated layout bounding boxes would inevitably occlude subjects. In particular, it should be noted that when the layout bounding box occludes the critical regions of the subject, such as the human head or face, the visual effect of the poster will be unpleasant, taking the image in row-3-column-6 as an example. In contrast, layout bounding boxes generated by PDA-GAN avoid subject regions nicely, thus the generated posters better express the information of subjects and layout elements.
 
 Meanwhile, the product occlusion degree $R_{sub}$ of PDA-GAN performance surpasses ContentGAN, Layoutprompter, and CGL-GAN by $39.8\%$, $18.1\%$, and $14.5\%$, respectively. The three rightmost columns in Fig.~\ref{fig:3} are the heat maps of the attention of each pixel to the product in the image. We get attention maps of product images (queried by their category tags extracted on product pages) by CLIP \cite{DBLP:conf/icml/RadfordKHRGASAM21,DBLP:conf/iccv/CheferGW21}. Compared with ContentGAN and CGL-GAN, PDA-GAN generates layout bounding boxes on the region with lower thermal values to avoid occluding products. For example, in the seventh column, the layout bounding box generated by PDA-GAN effectively avoids the region with high thermal values of the product, which enables the hoodie information of the product to be fully displayed.

 Compared to ContentGAN, LayoutPrompter, and CGL-GAN, PDA-GAN reduces the cFID score by $52.9\%$, $51.5\%$, and $19.1\%$, respectively, indicating better semantic harmony between layout and image content. Even in terms of the conventional FID, PDA-GAN achieves lower scores than the above baselines, demonstrating that it better captures the distribution of real poster layouts.
 The above quantitative and qualitative comparisons of models demonstrate that PDA-GAN improves the relationship modeling between graphic layouts and image contents, including color and texture details.

 To provide a more comprehensive evaluation beyond standard quantitative metrics, we conduct a user study, as summarized in Tab.~\ref{tab:user study}. A total of 120 test samples are randomly selected, each consisting of a product image and four corresponding layouts generated by ContentGAN, LayoutPrompter, CGL-GAN, and PDA-GAN. The participants include two groups: 10 professional designers and 20 novice designers. Each participant is asked to identify both the eligible and the best layout among the four candidates for each sample. We report the eligible selection rate ($P_e$) and best selection rate ($P_b$), defined as the proportion of votes received by each model relative to the total votes. Results show that PDA-GAN consistently outperforms other methods, particularly with a significantly higher proportion of best-selected ($P_b$) layouts.

\begin{table*}[!t]
    \caption{{\bf{Comprehensive comparison between CGL-GAN and PDA-GAN.}} \textnormal{Data \uppercase\expandafter{\romannumeral1} and Data \uppercase\expandafter{\romannumeral2} contain 8,000 and 54,546 source domain samples, respectively. \checkmark indicates the experiment configuration. The symbol "-" indicates that the model cannot complete the layout generation task since the generated element bounding boxes overlap with each other severely.}
    }
    \label{tab:CGL-GAN}
    \centering
    \setlength{\tabcolsep}{2.18mm}{
    \scalebox{1.16}{
    \begin{tabular}{lccc|ccc|cccc}
    \hline
         Model &Data \uppercase\expandafter{\romannumeral1} &Data \uppercase\expandafter{\romannumeral2} &Gaussian Blur  &$R_{com}\downarrow$ &$R_{shm}\downarrow$ &$R_{sub}\downarrow$ &$R_{ove}\downarrow$ &$R_{und}\uparrow$ &$R_{ali}\downarrow$ &$R_{occ}\uparrow$\\
    \hline
         CGL-GAN &\checkmark &\quad &\quad      &33.85 &13.88 &0.766 &0.0299 &0.9351 &0.0139 &\bf{99.7}\\
         CGL-GAN &\checkmark &\quad &\checkmark      &- &- &- &2.5826 &- &- &-\\

         CGL-GAN &\quad &\checkmark &\checkmark  &35.77 &15.47 &0.805 &\bf{0.0233} &0.9359 &\bf{0.0098} &99.6\\
         PDA-GAN &\checkmark &\quad &\checkmark      &36.41 &19.48 &1.044 &0.0244 &0.9384 &0.0091 &99.7\\

         PDA-GAN &\quad &\checkmark &\checkmark  &\bf{30.63} &18.14 &0.833 &0.0589 &0.9302 &0.0105 &99.6\\
         PDA-GAN &\checkmark &\quad &\quad   &33.55 &\bf{12.77} &\bf{0.688} &0.0290 &\bf{0.9481} &0.0105 &\bf{99.7}\\
    \hline
    \end{tabular}
    }
    }
\end{table*}

\begin{figure*}
    \centering
    \hspace{0cm}\includegraphics[width=\textwidth]{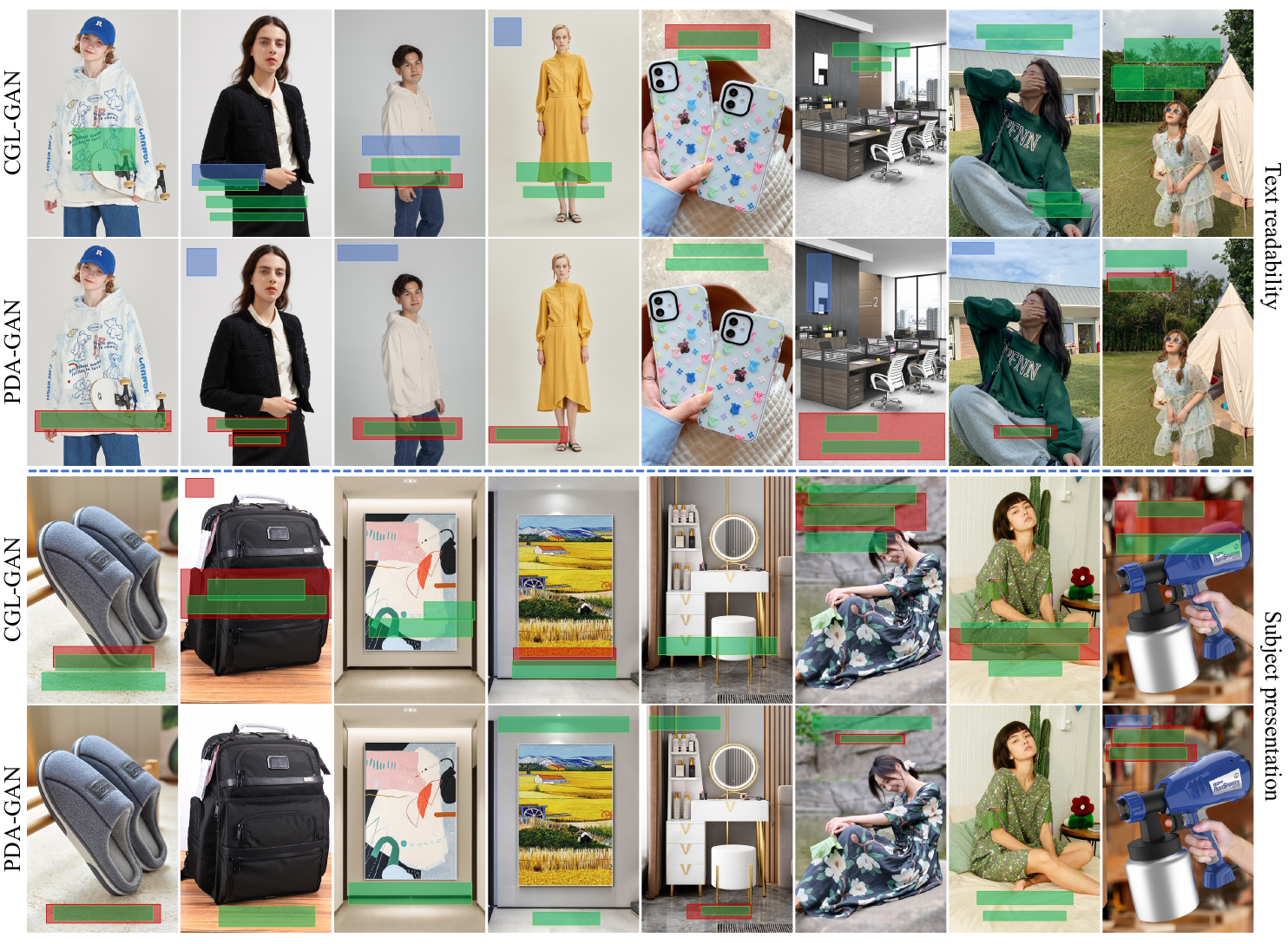}
    \caption{More qualitative comparisons between CGL-GAN and PDA-GAN.}
    \label{fig:CGL-PDA}
\end{figure*}

 \noindent\textbf{Layout generation without image contents.} We also compare our method with image-agnostic approaches, including LayoutTransformer~\cite{DBLP:conf/iccv/GuptaLA0MS21}, LayoutVTN~\cite{DBLP:conf/cvpr/ArroyoPT21}, LayoutDM~\cite{inoue2023layoutdm}, and LayoutDiffusion~\cite{zhang2023layoutdiffusion}. As shown in Tab.~\ref{tab:content-unaware}, these image-agnostic methods perform well on graphic metrics. However, our models significantly outperform them on content-aware metrics. Specifically, PDA-GAN surpasses LayoutTransformer, LayoutVTN, LayoutDM, and LayoutDiffusion by $18.0\%$, $19.7\%$, $18.6\%$, and $17.3\%$, respectively, in terms of background complexity $R_{com}$. This is because the image-agnostic methods only model the relationships between layout elements, without considering the image content.
These image-agnostic methods tend to generate bounding boxes for text elements in areas with complex backgrounds (as shown in the first two rows and the leftmost three columns of Fig.~\ref{fig:3}), which reduces the readability of the text information. Furthermore, compared with LayoutTransformer, LayoutVTN, LayoutDM, and LayoutDiffusion, PDA-GAN reduces $R_{shm}$ by $39.4\%$, $42.5\%$, $54.2\%$, and $53.3\%$, and reduces $R_{sub}$ by $47.5\%$, $48.0\%$, $61.6\%$, and $61.2\%$, respectively. The rightmost six columns in Fig.~\ref{fig:3} show that image-agnostic methods generate layout bounding boxes that randomly occlude subject and product areas. Such occlusions degrade the visibility and presentation of both subject and product content.

 \begin{figure*}[!t]
    \centering
    \hspace{0cm}\includegraphics[width=\textwidth]{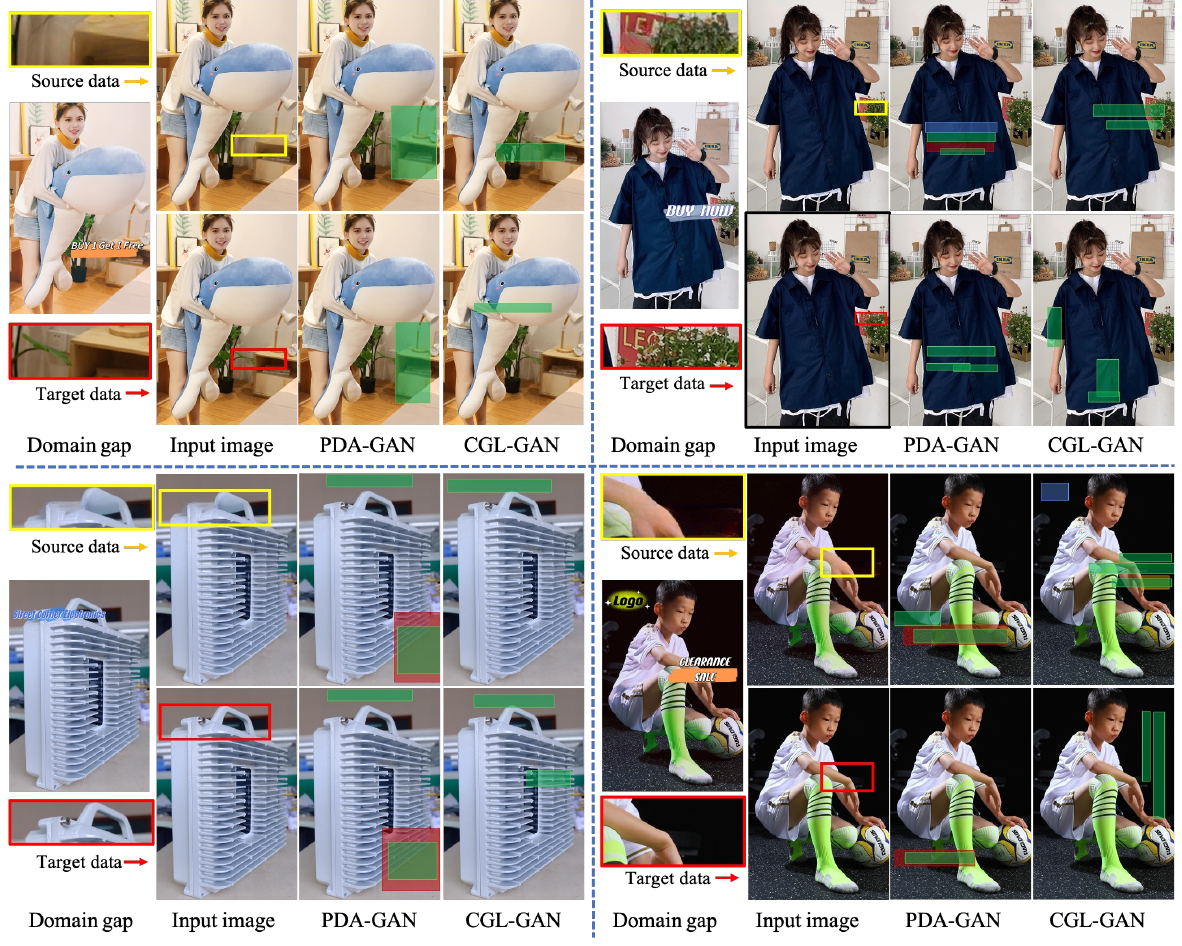}
    \caption{{\bf Layouts generated by different models using source and target domain data.} Inpainted images (source domain) and clean images (target domain) are both fed into PDA-GAN and CGL-GAN. The results produced by PDA-GAN are relatively consistent, indicating that our method achieves better feature alignment between the two domains.}
    \label{fig:Eliminating the domain gap}
\end{figure*}

 \noindent\textbf{More comparisons with CGL-GAN.} 
 As shown in the first and last rows of Tab.~\ref{tab:CGL-GAN}, PDA-GAN consistently outperforms CGL-GAN across all metrics under identical configurations, validating the effectiveness of the proposed PD. Unlike CGL-GAN, PDA-GAN replaces both the Gaussian blur pre-processing and the heavy discriminator network with a lightweight PD module (only 332,545 parameters, less than $2\%$ of the 22,575,841 parameters in CGL-GAN), significantly reducing memory and computation cost.
To enable a fair comparison, PDA-GAN is also trained under the same configuration as CGL-GAN. As shown in the middle rows of Tab.~\ref{tab:CGL-GAN}, CGL-GAN trained on Data~\uppercase\expandafter{\romannumeral1} with Gaussian blur fails to produce usable layouts, resulting in extremely high $R_{ove}$ due to severely overlapping layout elements. In contrast, PDA-GAN under the same setting generates reasonable and well-structured layouts. These results highlight the stronger generalization ability of PDA-GAN across different domain conditions.
Intuitively, Gaussian blur can help narrow the domain gap, but it also causes loss of image color and texture details. As shown in the fourth and sixth rows of Tab.~\ref{tab:CGL-GAN}, PDA-GAN without Gaussian blur achieves the best performance across all content-aware metrics. This result shows that PDA-GAN can eliminate the domain gap without relying on Gaussian blur, while preserving image details for generating high-quality layouts.
 
 As illustrated in the sixth and eighth columns of the first two rows in Fig.~\ref{fig:CGL-PDA}, compared with CGL-GAN, PDA-GAN generates text bounding boxes with a simpler background. It is interesting to observe from the first two rows of Fig.~\ref{fig:CGL-PDA} that when PDA-GAN generates boxes among complex backgrounds, it tends to additionally generate an underlay bounding box which covers the complex background to ensure readability of the text information. The last two rows show that layouts generated by PDA-GAN can effectively avoid the subject area, and then can generate posters that better express the information of subjects and layout elements.
 Both the above quantitative and qualitative evaluations demonstrate that PDA-GAN can capture the subtle interaction between image contents and graphic layouts and achieve the SOTA performance.

\subsection{Eliminating Domain Gap} \label{Eliminating DG}
\noindent As shown in Fig.~\ref{fig:Eliminating the domain gap}, we randomly select four clean product images ${\boldsymbol{x}_{img}}$ from the target domain data and add graphic layout elements to these images to create advertising posters ${\boldsymbol{x}_{pst}}$. Inpainting the regions of elements in posters to obtain inpainted images ${\boldsymbol{x}^{inp}_{pst}}$. Due to inpainted areas, there is a domain gap between ${\boldsymbol{x}_{img}}$ (target domain) and ${\boldsymbol{x}^{inp}_{pst}}$ (source domain).
 
To demonstrate that PDA-GAN can effectively bridge the domain gap, we input ${\boldsymbol{x}^{inp}_{pst}}$ and ${\boldsymbol{x}_{img}}$ to CGL-GAN and PDA-GAN to generate layouts. The mean difference values of the shallow-level feature maps, fusion feature maps, and deep-level feature maps generated by CGL-GAN between ${\boldsymbol{x}^{inp}_{pst}}$ and ${\boldsymbol{x}_{img}}$ of the above four samples as input are 0.0610, 0.1289, and 0.1263. The corresponding mean values calculated by PDA-GAN are 0.0293, 0.0726, and 0.1160, respectively. Compared with CGL-GAN, PDA-GAN has less difference in the generated feature maps between the source and target domain data.

From the perspective of generated results, layouts generated by CGL-GAN under different input domains (i.e., clean images ${\boldsymbol{x}_{img}}$ vs. inpainted images ${\boldsymbol{x}^{inp}_{pst}}$) exhibit noticeable inconsistencies. Specifically, CGL-GAN tends to place layout elements in distorted or blurred regions of the inpainted images, likely because the annotated bounding boxes in the source domain have all been inpainted during training. In contrast, PDA-GAN produces layouts that are visually more consistent across these two inputs. To validate the effectiveness of the pixel-level discriminator in bridging the domain gap, we further compute FID between the layouts generated from ${\boldsymbol{x}_{img}}$ and ${\boldsymbol{x}^{inp}_{pst}}$. PDA-GAN achieves a significantly lower FID score of 12.67 compared to 15.34 for CGL-GAN, indicating enhanced robustness to domain shifts. This result, along with the feature-level comparison, demonstrates that PDA-GAN can effectively eliminate the domain gap caused by inpainting.

 \subsection{Ablations}
 \noindent\textbf{Effects of the discriminator at the pixel level.} We first compare our PD with a global discriminator (GD) that only predicts one real or fake probability as in classical GAN. The abbreviation GD in Tab.~\ref{tab:PDA-DA} indicates the global discriminator strategy. When the weight of GD loss ($\gamma$ in Eq.~\ref{eq3}) is more than 0.01, the model cannot complete the layout generation task, indicated by the symbol $-$, since the $R_{ove}$ value is too high. From the statistics in Tab.~\ref{tab:PDA-DA}, our PD outperforms GD on all metrics. 
 
Second, we compare PD with the PatchGAN strategy~\cite{DBLP:conf/cvpr/IsolaZZE17}. 
The patch size in the Tab.~\ref{tab:PDA-PatchDA} refers to the dimension of the output map, which is then compared with the correspondingly resized ground-truth white-patch map during training. We train these models with $\gamma$ in Eq.~\ref{eq3} set to 6. The values of quantitative metrics listed in Tab.~\ref{tab:PDA-PatchDA} also confirm the advantage of the pixel-level discriminator. These experiments demonstrate that, due to the discrepancy by inpainting at the pixel level, the model might need to eliminate the domain gap at the pixel level. Additionally, the pixel-level strategy in the last row of Tab.~\ref{tab:PDA-PatchDA} can be considered to be the most fine-grained approach at the patch level.

 \begin{table}[!t]
    \caption{{\bf{Ablation study with global discriminators.}} \textnormal{$W$ refers to the weight of GD (or PD) module loss in the training process.}}
    \label{tab:PDA-DA}
    \centering
    \setlength{\tabcolsep}{0.58mm}{
    \scalebox{1.16}{
    \begin{tabular}{l|ccc|cccc}
    \hline
         Model-$W$  &$R_{com}\downarrow$ &$R_{shm}\downarrow$ &$R_{sub}\downarrow$ &$R_{ove}\downarrow$ &$R_{und}\uparrow$ &$R_{ali}\downarrow$\\
    \hline
         GD-6.0      &- &- &- &9.0000 &- &-\\
         GD-1.0      &- &- &- &8.9995 &- &-\\
         GD-0.01     &- &- &- &4.7764 &- &-\\
         GD-0.001      &34.41 &13.78 &0.749 &0.0327 &0.9299 &0.0110\\
         GD-0.0001     &34.77 &14.62 &0.777 &0.0345 &0.9234 &0.0122\\
         GD-0.0     &34.07 &15.13 &0.800  &0.0350 &0.9259 &0.0108\\

         PD-6.0   &\bf{33.55} &\bf{12.77} &\bf{0.688} &\bf{0.0290} &\bf{0.9481} &\bf{0.0105}\\
    \hline
    \end{tabular}
    }
    }
\end{table}

\begin{table}[!t]
    \caption{{\bf{Ablation study with PatchGAN-based methods.}} \textnormal{The input image height and width are 320 and 240 respectively.}}
    \label{tab:PDA-PatchDA}
    \centering
    \setlength{\tabcolsep}{0.58mm}{
    \scalebox{1.16}{
    \begin{tabular}{l|ccc|cccc}
    \hline
         Patch size  &$R_{com}\downarrow$ &$R_{shm}\downarrow$ &$R_{sub}\downarrow$ &$R_{ove}\downarrow$ &$R_{und}\uparrow$ &$R_{ali}\downarrow$\\
    \hline
         12*8      &- &- &- &0.9288 &- &-\\
         24*16      &33.67 &16.00 &0.844 &0.0438 &0.9407 &\bf{0.0075}\\
         44*30      &34.03 &13.02 &0.752 &\bf{0.0284} &0.9377 &0.0119\\
         88*60      &\bf{32.65} &13.35 &0.735 &0.0325 &0.9173 &0.0094\\

         350*240   &33.55 &\bf{12.77} &\bf{0.688} &0.0290 &\bf{0.9481} &0.0105\\
    \hline
    \end{tabular}
    }
    }
\end{table}

 \noindent\textbf{Effects of PD with different level feature maps.} In our model, PD is connected to the shallow-level feature maps of the first residual block. We now investigate PD performance when utilizing the deep-level feature from the fourth residual block and the fused feature (fusion of feature maps from the first to fourth residual blocks) in multi-scale CNN. As shown in Tab.~\ref{tab:PDA_layer}, discriminating with shallow feature maps in PDA-GAN can achieve better results in both content-aware and graphic metrics on average. These results further validate the effectiveness of our PD design. Intuitively, bridging the domain gap at an early stage in the network may benefit subsequent processing within the model.

 \noindent\textbf{Effects of PD with different architectures.} To validate the robustness and adaptability of the proposed PD, we conduct ablation studies focusing on two architectural aspects: kernel size and network depth. First, we compare PD modules using convolutional kernels of size 3×3, 5×5, 7×7, as well as a hybrid configuration that stacks layers with 3, 5, and 7-sized kernels. As shown in Tab.~\ref{tab:PD_size}, the performance remains stable across different configurations, with the 3×3 kernel achieving slightly better results on most metrics. 
In addition, we investigate the impact of network depth by varying the number of transposed convolution layers in PD (2, 3, 6, and 9 layers). The results, summarized in Tab.~\ref{tab:PD_layers}, indicate that performance is comparable across different depths. These results confirm that the proposed PD design is compact and effective, with a lightweight 3×3 kernel and a relatively shallow network being sufficient to handle the domain discrepancies introduced by inpainting.

 \noindent\textbf{Effects of label smoothing.}  
 For the ground truth white-patch map input to the discriminator, the two-side label smoothing means we set 0 to 0.2 and 1 to 0.8, the one-source label smoothing means we only set 1 to 0.8, and the one-target label smoothing means we only set 0 to 0.2. The first row in Tab.~\ref{tab:label_smoothing} means the model without label smoothing. Tab.~\ref{tab:label_smoothing} shows that the model with one-target label smoothing performs better in all metrics than without label smoothing, demonstrating its effectiveness. In addition, the effects of two-side or one-source label smoothing are not as good as one-target label smoothing on average. 
 In contrast, both two-side and one-source label smoothing perform worse than one-target label smoothing.
 
\begin{table}[!t]
    \caption{{\bf{Quantitative ablation study on different level feature maps for the pixel-level discriminator.}}}
    \label{tab:PDA_layer}
    \centering
    \setlength{\tabcolsep}{0.54mm}{
    \scalebox{1.16}{
    \begin{tabular}{lccc|ccc}
    \hline
         Feature map   &$R_{com}\downarrow$ &$R_{shm}\downarrow$ &$R_{sub}\downarrow$ &$R_{ove}\downarrow$ &$R_{und}\uparrow$ &$R_{ali}\downarrow$\\
    \hline
         deep level      &34.22 &13.97 &0.770 &0.0396 &0.9366 &0.0118\\
         fusion      &35.36 &14.54 &0.817 &0.0310 &\bf{0.9513} &0.0117\\
         shallow level     &\bf{33.55} &\bf{12.77} &\bf{0.688} &\bf{0.0290} &0.9481 &\bf{0.0105}\\
    \hline
    \end{tabular}}
    }
\end{table}

\begin{table}[!t]
    \caption{{\bf{Quantitative ablation study on PD with different kernel sizes.}} \textnormal{3, 5, and 7 denote convolutional kernel sizes of 3×3, 5×5, and 7×7, respectively, while 'fusion' indicates a hybrid design combining all three.}}
    \label{tab:PD_size}
    \centering
    \setlength{\tabcolsep}{0.54mm}{
    \scalebox{1.16}{
    \begin{tabular}{lccc|ccc}
    \hline
         Size   &$R_{com}\downarrow$ &$R_{shm}\downarrow$ &$R_{sub}\downarrow$ &$R_{ove}\downarrow$ &$R_{und}\uparrow$ &$R_{ali}\downarrow$\\
    \hline
         5      &33.09 &13.66 &0.720 &0.0349 &0.8891 &0.0086\\
         7      &\bf{31.69} &14.73 &0.743 &0.0431 &0.9185 &\bf{0.0081}\\
        fusion     &32.30 &13.25 &0.747 &0.0333 &0.9316 &0.0095\\
         3     &33.55 &\bf{12.77} &\bf{0.688} &\bf{0.0290} &\bf{0.9481} &0.0105\\
    \hline
    \end{tabular}}
    }
\end{table}

\begin{table}[!t]
    \caption{{\bf{Quantitative ablation study on different level feature maps for the pixel-level discriminator.}} \textnormal{2, 3, 6, and 9 denote the number of transposed convolutional layers used in the PD module.}}
    \label{tab:PD_layers}
    \centering
    \setlength{\tabcolsep}{0.54mm}{
    \scalebox{1.16}{
    \begin{tabular}{lccc|ccc}
    \hline
         Layers   &$R_{com}\downarrow$ &$R_{shm}\downarrow$ &$R_{sub}\downarrow$ &$R_{ove}\downarrow$ &$R_{und}\uparrow$ &$R_{ali}\downarrow$\\
    \hline
         2      &\bf{32.37} &13.48 &\bf{0.687} &0.0360 &0.9151 &0.0110\\
         6      &32.97 &15.25 &0.792 &0.0318 &0.9040 &0.0130\\
         9     &34.07 &13.80 &0.743 &\bf{0.0279} &0.9295 &\bf{0.0086}\\
         3     &33.55 &\bf{12.77} &0.688 &0.0290 &\bf{0.9481} &0.0105\\
    \hline
    \end{tabular}}
    }
\end{table}

\begin{table}[!t]
    \caption{{\bf{Ablation study with different label smoothing choices.}} \textnormal{The first row is the model without label smoothing. Two-side: set 0 to 0.2 and 1 to 0.8; one-source: set 1 to 0.8; and 0ne-target: set 0 to 0.2.}}
    \label{tab:label_smoothing}
    \centering
    \setlength{\tabcolsep}{0.6mm}{
    \scalebox{1.16}{
    \begin{tabular}{lccc|ccc}
    \hline
         smoothing   &$R_{com}\downarrow$ &$R_{shm}\downarrow$ &$R_{sub}\downarrow$ &$R_{ove}\downarrow$ &$R_{und}\uparrow$ &$R_{ali}\downarrow$\\
    \hline
         without      &33.61 &14.04 &0.718 &0.0346 &0.9188 &0.0106\\
         two-side      &33.66 &14.67 &0.794 &0.0334 &0.9297 &0.0098\\
         one-source    &\bf{32.20} &15.23 &0.799 &0.0431 &0.9234 &\bf{0.0085}\\
         one-target     &33.55 &\bf{12.77} &\bf{0.688} &\bf{0.0290} &\bf{0.9481} &0.0105\\
    \hline
    \end{tabular}}
    }
\end{table}

\noindent\textbf{Effects of PD.} Compared to the model without the PD module in the first row of Tab.~\ref{tab:PDA-DA}, under the same configuration, the model with the PD module achieves better results in all metrics. Benefiting from the PD module, which effectively eliminates the domain gap, as demonstrated in Sec.~\ref{Eliminating DG}, the model with the PD module can generate high-quality, image-aware graphic layouts for advertising posters.

\noindent\textbf{Replacing DETR with deformable DETR.} We further evaluate the generalization ability of our approach by replacing the DETR-based layout generator with deformable DETR, a multi-scale transformer model known for its better convergence and object localization. As shown in Tab.~\ref{tab:DDETR}, our method maintains strong performance, with DETR yielding slightly better results on several layout accuracy metrics. Since the PD module performs domain adaptation on shallow visual features, feeding only high-level features into the transformer can be more effective, as the domain gap has already been eliminated at earlier stages. These results validate that our method generalizes well across different detection backbones and that the PD module remains effective with both DETR and deformable DETR.

\begin{figure*}[!t]
    \centering
    \hspace{0cm}\includegraphics[width=17.6cm]{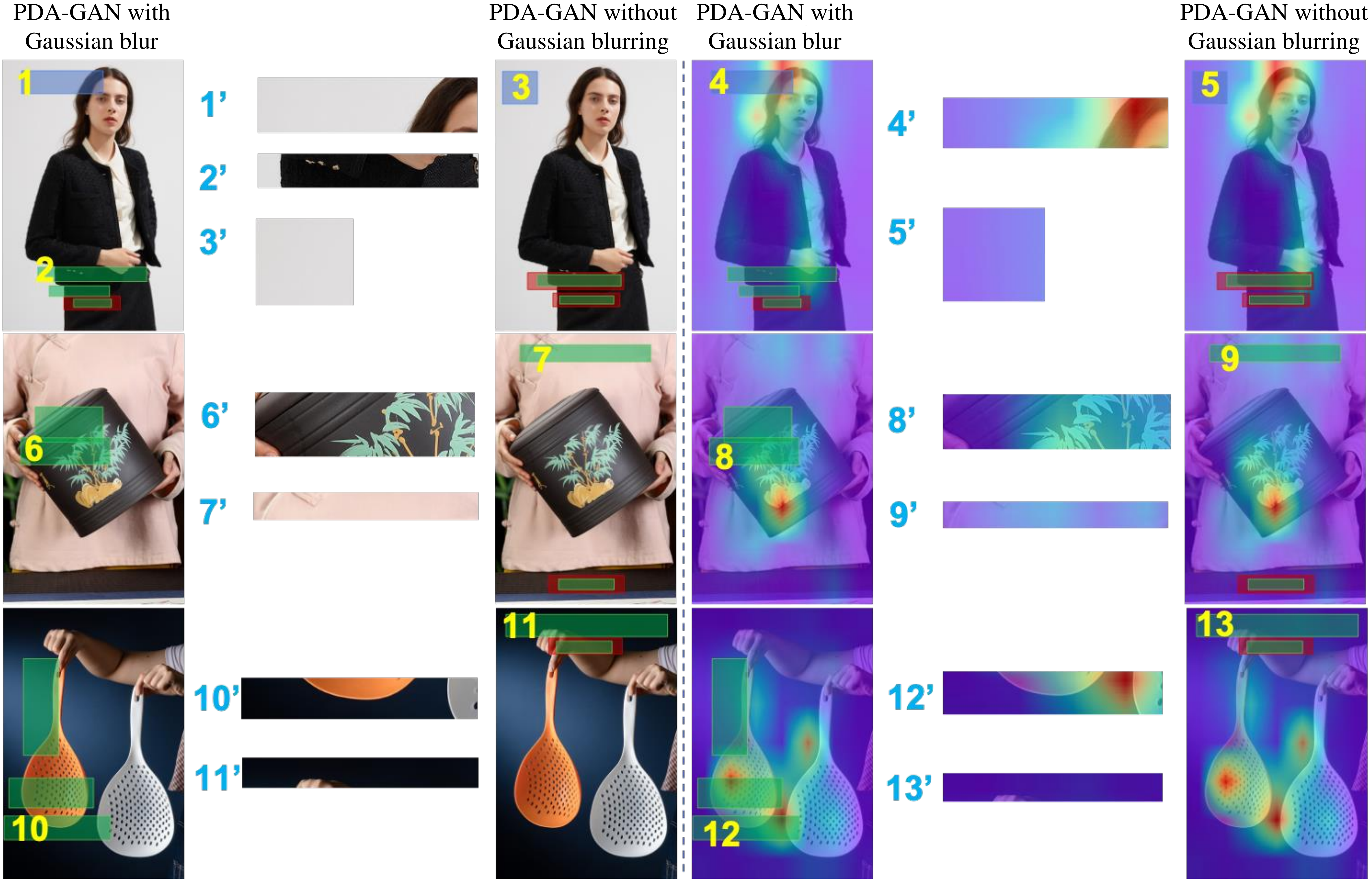}
    \caption{{\bf Impact of Gaussian blur.} Layouts in each row are generated using the same input image, while those in each column correspond to different input samples. “PDA-GAN with Gaussian blur” indicates that the input data is preprocessed with Gaussian blurring. The blue-numbered boxes in the middle show enlarged views of the regions marked with yellow numbers. The left part of the vertical dotted line displays the input images, while the right part displays the corresponding product attention heatmaps.}
    \label{fig:4}
\end{figure*}

\begin{table}[!t]
    \caption{{\bf{Quantitative ablation study on PD.}} \textnormal{Ours' refers to our model of PDA-GAN without the PD module.}}
    \label{tab:PDA-DA}
    \centering
    \setlength{\tabcolsep}{0.62mm}{
    \scalebox{1.16}{
    \begin{tabular}{lccc|cccc}
    \hline
         Model  &$R_{com}\downarrow$ &$R_{shm}\downarrow$ &$R_{sub}\downarrow$ &$R_{ove}\downarrow$ &$R_{und}\uparrow$ &$R_{ali}\downarrow$\\
    \hline
         Ours'     &34.07 &15.13 &0.800  &0.0350 &0.9259 &0.0108\\
         Ours      &\bf{33.55} &\bf{12.77} &\bf{0.688} &\bf{0.0290} &\bf{0.9481} &\bf{0.0105}\\
    \hline
    \end{tabular}
    }
    }
\end{table}

\begin{table}[!t]
    \caption{{\bf{Quantitative ablation study on different backbones.}} \textnormal{D-DETR denotes deformable DETR.}}
    \label{tab:DDETR}
    \centering
    \setlength{\tabcolsep}{0.54mm}{
    \scalebox{1.16}{
    \begin{tabular}{lccc|ccc}
    \hline
         Backbone   &$R_{com}\downarrow$ &$R_{shm}\downarrow$ &$R_{sub}\downarrow$ &$R_{ove}\downarrow$ &$R_{und}\uparrow$ &$R_{ali}\downarrow$\\
    \hline
         D-DETR     &\bf{32.48} &13.03 &0.698 &0.1085 &0.9423 &\bf{0.0078}\\
         DETR     &33.55 &\bf{12.77} &\bf{0.688} &\bf{0.0290} &\bf{0.9481} &0.0105\\
    \hline
    \end{tabular}}
    }
\end{table}

 \begin{table}[!t]
    \caption{{\bf{Quantitative ablation study on Gaussian blur.}} \textnormal{Ours* refers to the PDA-GAN model with Gaussian blur applied to the input image.}}
    \label{tab:PDA+blur}
    \centering
    \setlength{\tabcolsep}{0.62mm}{
    \scalebox{1.16}{
    \begin{tabular}{lccc|ccc}
    \hline
         Model   &$R_{com}\downarrow$ &$R_{shm}\downarrow$ &$R_{sub}\downarrow$ &$R_{ove}\downarrow$ &$R_{und}\uparrow$ &$R_{ali}\downarrow$\\
    \hline
         Ours*      &36.71 &20.14 &1.036 &0.0475 &0.9376 &\bf{0.0068}\\
         Ours     &\bf{33.55} &\bf{12.77} &\bf{0.688} &\bf{0.0290} &\bf{0.9481} &0.0105\\
    \hline
    \end{tabular}}}
\end{table}
 
 \noindent\textbf{Effect of Gaussian blur.} Based on the PDA-GAN model, we quantitatively and qualitatively analyze the impact of Gaussian blur on layout generation, as shown in Tab.~\ref{tab:PDA+blur} and Fig.~\ref{fig:4}. Applying Gaussian blur leads to an increase in $R_{com}$ from 33.55 to 36.71. As shown in boxes 2, 6, and 10 of Fig.~\ref{fig:4}, the model with Gaussian blur tends to generate text bounding boxes with more complex backgrounds, which reduces text readability. In contrast, boxes 7 and 11 illustrate that the model without Gaussian blur generates text bounding boxes with simpler backgrounds or introduces underlay bounding boxes to replace complex backgrounds.
 
 Tab.~\ref{tab:PDA+blur} shows that $R_{shm}$ and $R_{sub}$ of the model with Gaussian blur increase from 12.77 to 20.14 and from 0.688 to 1.036, respectively. As shown in Fig.~\ref{fig:4}, the model with Gaussian blur tends to generate layout bounding boxes that occlude subject or product regions. Such layouts diminish the visibility of subjects and the clarity of layout elements in advertising posters. These quantitative and qualitative results demonstrate that the loss of image details caused by Gaussian blur degrades the quality of the generated image-aware graphic layouts.

\begin{figure*}[!t]
    \centering
    \hspace{0cm}\includegraphics[width=17.6cm]{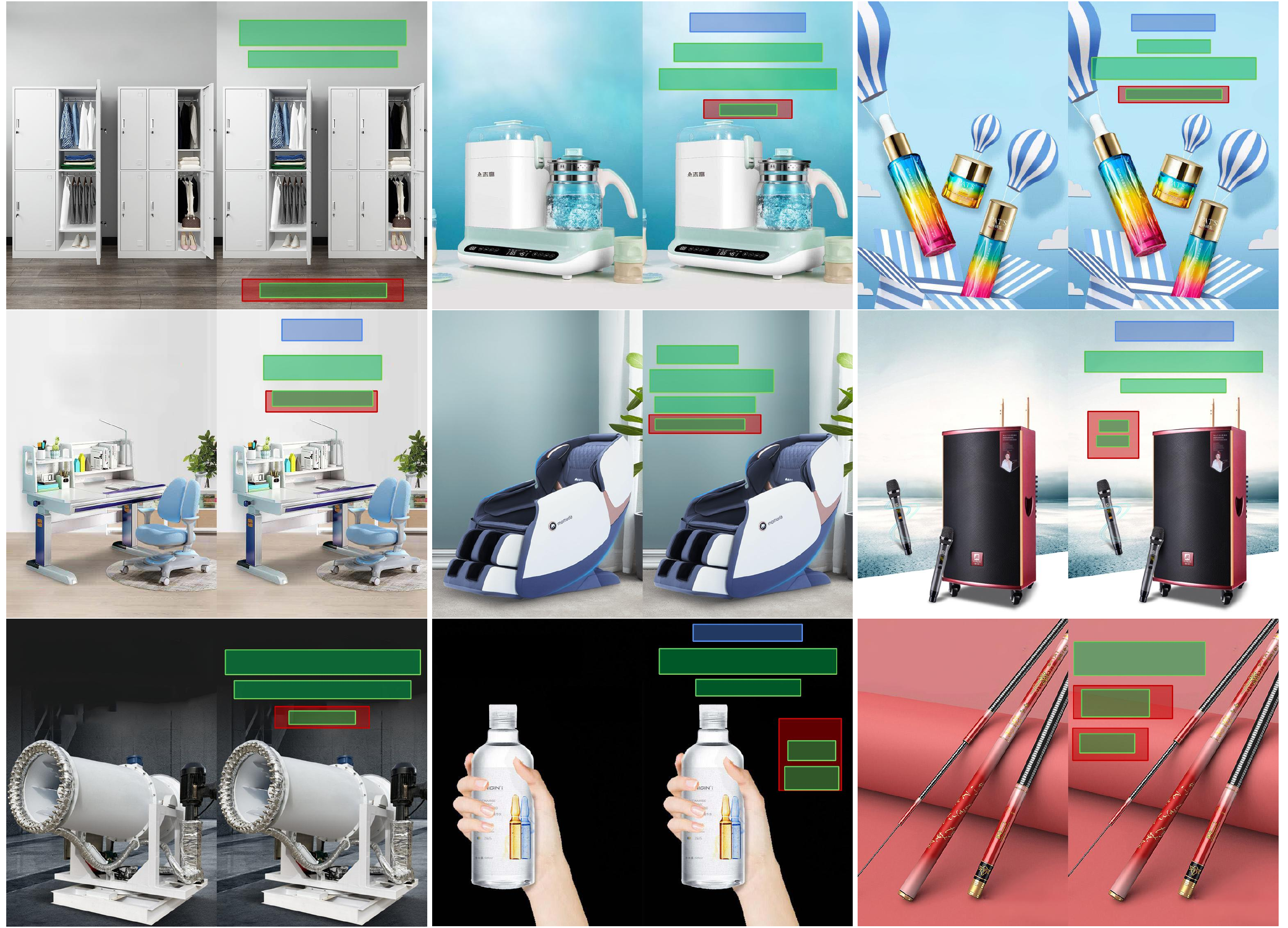}
    \caption{{\bf Generalization ability of PDA-GAN on PKU-Dataset~\cite{DBLP:conf/cvpr/HsuHPKZ23}.}}
    \label{fig:PDA_PLD}
\end{figure*}

\subsection{Extension Experiments}

\noindent\textbf{Effect of training dataset.} To further validate the advantages of the CGL-Dataset for image-aware layout generation, we conduct additional experiments using the publicly available PKU-Dataset~\cite{DBLP:conf/cvpr/HsuHPKZ23}, which, as introduced in Sec.~\ref{Datasets}, includes a broader variety of poster types and layout styles. Specifically, we train the PDA-GAN model separately on the CGL-Dataset and the PKU-Dataset, and evaluate both models on 1,000 clean product images. As shown in Tab.~\ref{tab:Datasets}, the PDA-GAN trained on the CGL-Dataset consistently outperforms the model trained on the PKU-Dataset across all evaluation metrics. This result highlights the benefits of the CGL-Dataset, which, compared to the PKU-Dataset, offers a larger scale, more diverse element categories, and better alignment with the needs of image-aware layout generation.

 \begin{table}[!t]
    \caption{{\bf{Comparison of PDA-GAN trained on different datasets.}}}
    \label{tab:Datasets}
    \centering
    \setlength{\tabcolsep}{0.62mm}{
    \scalebox{1.16}{
    \begin{tabular}{lccc|ccc}
    \hline
         Dataset   &$R_{com}\downarrow$ &$R_{shm}\downarrow$ &$R_{sub}\downarrow$ &$R_{ove}\downarrow$ &$R_{und}\uparrow$ &$R_{ali}\downarrow$\\
    \hline
         PKU~\cite{DBLP:conf/cvpr/HsuHPKZ23}      &33.88 &18.03 &0.881 &0.0445 &0.9411 &0.0127\\
         CGL(ours)     &\bf{33.55} &\bf{12.77} &\bf{0.688} &\bf{0.0290} &\bf{0.9481} &{0.0105}\\
    \hline
    \end{tabular}}}
\end{table}

\noindent\textbf{Generality to other poster types.} To further demonstrate the generalizability of the PDA-GAN, we evaluate the model's performance on the PKU-Dataset test set~\cite{DBLP:conf/cvpr/HsuHPKZ23}, as shown in Fig.~\ref{fig:PDA_PLD}. This dataset contains a diverse range of poster types with varying design styles and layout structures. The generated layouts effectively avoid interference with the main subject areas of the posters, ensuring that both the background image and layout elements are displayed harmoniously. This demonstrates that PDA-GAN can successfully generalize to new poster types and adapt to different layout styles, further validating its robustness and flexibility.

\begin{figure}[ht]
    \centering
    \hspace{-0.2cm}\includegraphics[width=8.8cm]{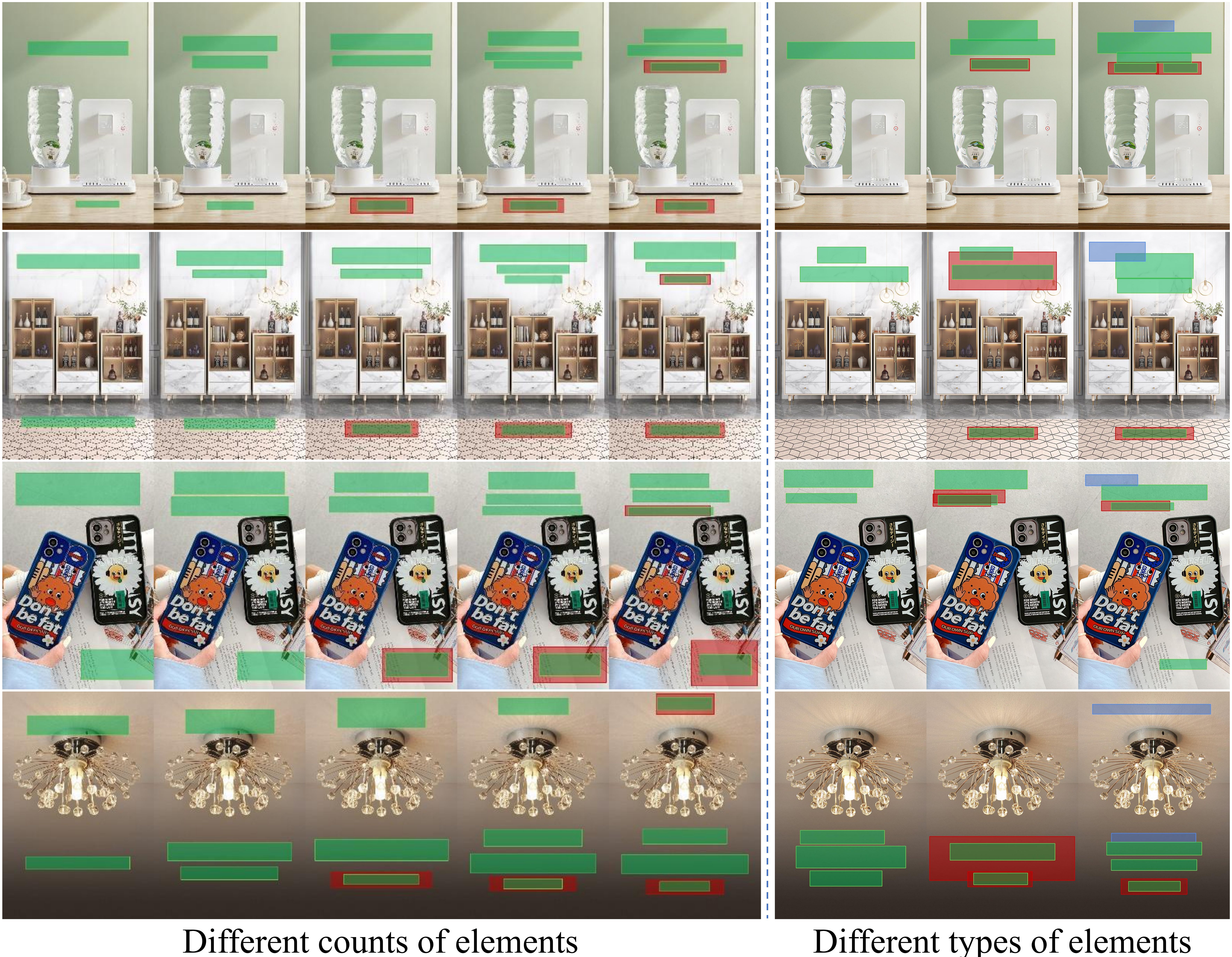}
    \vspace{0.0cm}
    \caption{{\bf Language-guided layout generation with varying element counts and types.} Each row corresponds to the same background image. The left part shows layouts generated from prompts specifying the number of elements (e.g., ``generate three elements", 2 to 6 from left to right), while the right part shows layouts guided by prompts specifying element types (e.g., ``generate text and underlay").}
    \label{fig:GPT_CLIP_PD}
\end{figure}

\noindent\textbf{Extension to natural language-guided layout generation with LLMs.} 
Although originally designed as a discriminator in a GAN-based pipeline, the PD module can be flexibly integrated with other types of generation backbones, including LLM-based and diffusion-based models. To explore this adaptability, we extend our framework to a language-guided layout generation setting, where the PD module is integrated with LLM-based components. In this extension, CLIP~\cite{DBLP:conf/icml/RadfordKHRGASAM21} is used to extract textual features from user-provided natural language prompts, which are then fused with multi-scale image features. These combined features are passed through a transformer encoder, followed by fully connected layers, to generate layout predictions. The input language prompts are structured as simple declarative sentences, such as ``generate three elements" or ``generate text and underlay". To enhance generalization, we use GPT~\cite{brown2020language} to augment the input prompts, generating multiple semantically similar variants during training. This allows the model to better capture the diversity of natural language expressions. As illustrated in Fig.~\ref{fig:GPT_CLIP_PD}, the model is capable of generating image-aware layouts that are consistent with both the semantic content of the image and a broad range of natural language instructions, including those that specify the number and categories of layout elements. These results demonstrate that the proposed PD module is a flexible component that can be adapted to a variety of generation frameworks, including those guided by natural language.

\begin{figure}[ht]
    \centering
    \hspace{-0.2cm}\includegraphics[width=8.8cm]{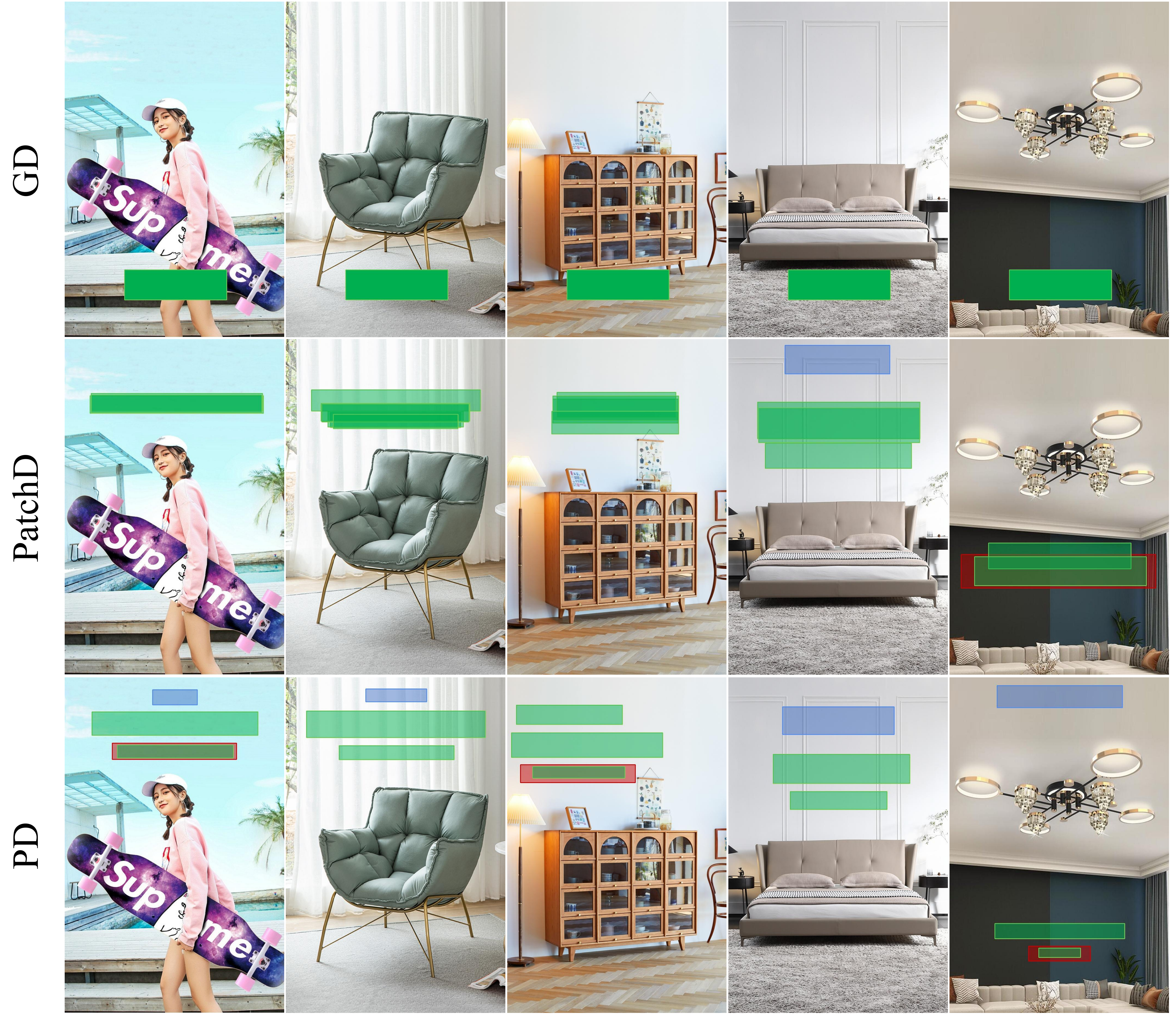}
    \vspace{0.0cm}
    \caption{{\bf Failure cases.} The first to third rows show layouts generated by models with the global discriminator (GD), patch discriminator (PatchD), and our PD module, respectively.}
    \label{fig:failure_case}
\vspace{0cm}
\end{figure}

\noindent\textbf{Analysis of failure cases from alternative discriminator designs.} To complement our evaluation, we analyze the failure cases of alternative discriminator designs, specifically those using either global or patch-level supervision, trained with identical configurations as our full model. Quantitative results are presented in the first rows of Tab.~\ref{tab:PDA-DA} and Tab.~\ref{tab:PDA-PatchDA}, where both alternatives yield significantly higher layout overlap ratios. As shown in the first and second rows of Fig.~\ref{fig:failure_case}, the generated layouts exhibit severe element overlap and visual clutter, making them unsuitable for practical poster design applications. These examples demonstrate that directly applying global or patch-level discriminators fails to effectively address the domain gap, and thus cannot produce usable image-aware layouts. In contrast, our pixel-level discriminator enables fine-grained domain alignment at the early visual stage, resulting in high-quality layouts that are better aligned with image content.

\section{Conclusion}
\vspace{0.0cm}
In this paper, we focus on generating image-aware graphic layouts for advertising posters. We introduce a novel generative framework, PDA-GAN, designed to first bridge the domain gap and then model the relationship between image content and layouts. For the image-aware layout generation task, we contribute a large advertising graphic layout dataset and propose three novel content-aware metrics. Both quantitative and qualitative evaluations demonstrate that our method achieves state-of-the-art performance and generates high-quality image-aware graphic layouts for posters. To support further research in graphic layout generation, we will release our model code and dataset to the community. In the future, we plan to explore how to better integrate user constraints, such as element categories and coordinates, and enhance layout generation diversity. Additionally, we aim to develop an automated system for end-to-end generation of high-quality posters directly from product images.

\vspace{0.0cm}
\section*{Acknowledgments}
\vspace{0.0cm}
We sincerely thank the anonymous reviewers for their professional and constructive comments, which have helped us improve the quality and clarity of this paper. Weiwei Xu is partially supported by “Pioneer” and “Leading Goose” R\&D Program of Zhejiang (No. 2023C01181). This work is supported by Alibaba Group through the Alibaba Innovation Research Program, the State Key Lab of CAD\&CG, and the Information Technology Center, Zhejiang University.

\bibliographystyle{IEEEtran}
\bibliography{cite}

\begin{IEEEbiography}[{\includegraphics[width=1in,height=1.25in,clip,keepaspectratio]{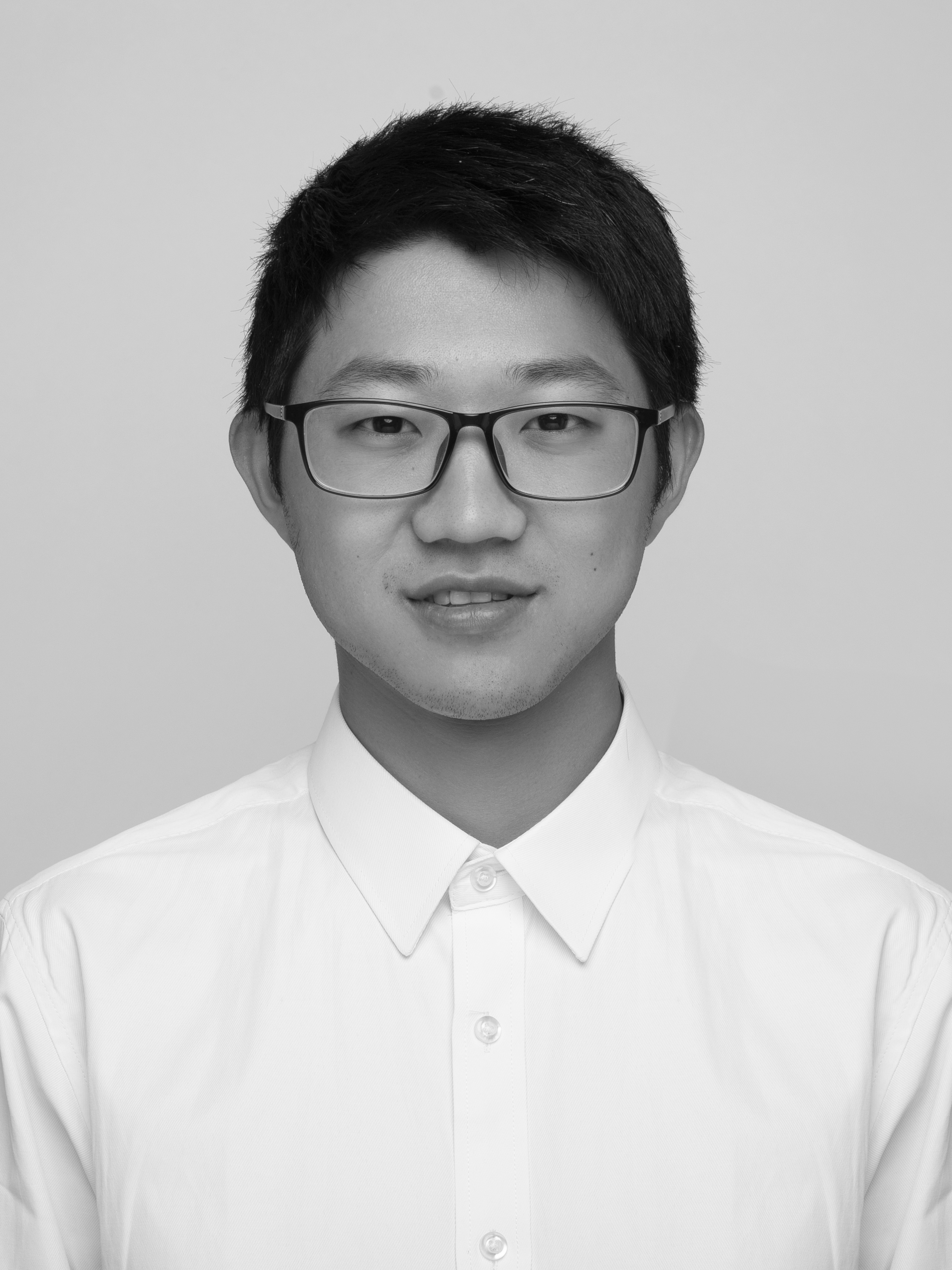}}]{Chenchen Xu}
received the B.Sc. and M.Sc. degrees from Anhui Normal University, Wuhu, China, in 2016 and 2020, respectively. He received the Ph.D. degree from the State Key Lab of CAD \& CG, Zhejiang University, Hangzhou, China, in 2024, under the supervision of Prof. Weiwei Xu. He is currently a researcher at Anhui Normal University and The Chinese University of Hong Kong. His research interests include image processing and machine learning, with a focus on deep learning, graphic layout generation, and image matting.
\end{IEEEbiography}

\begin{IEEEbiography}[{\includegraphics[width=1in,height=1.25in,clip,keepaspectratio]{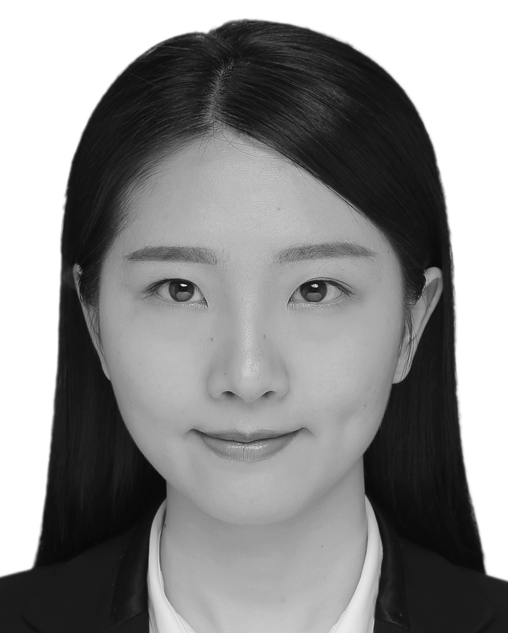}}]{Min Zhou}
received the B.S. and M.S. degrees from Beihang University, Beijing, China, in 2016 and 2019, respectively. She is currently a researcher in Alibaba Group. Her research interests include computer vision and deep learning.
\end{IEEEbiography}

\begin{IEEEbiography}[{\includegraphics[width=1in,height=1.25in,clip,keepaspectratio]{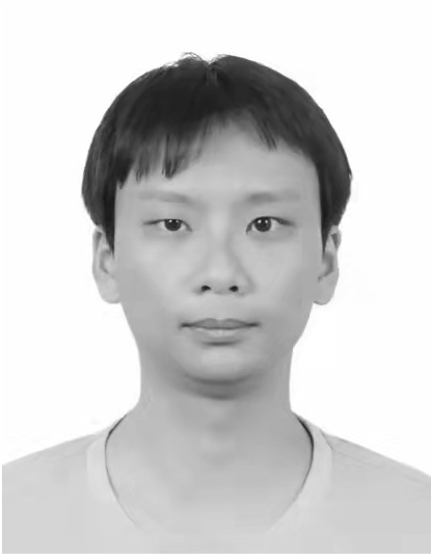}}]{Tiezheng Ge}
received his B.S. and Ph.D. degree from University of Science and Technology of China in 2009 and 2014 respectively. After that, he joined Alibaba Group. Now, he serves as a Staff Algorithm Engineer in Alimama(the Advertising Department of Alibaba), leading a research group of intellegent ad creative designing. His recent research interest includes the smart generation of image/video/text for online e-Commercial product, and relevant technics such as motion transfer, image matting, image/video caption, and image inpainting.
\end{IEEEbiography}

\vspace{-15cm}

\begin{IEEEbiography}[{\includegraphics[width=1in,height=1.25in,clip,keepaspectratio]{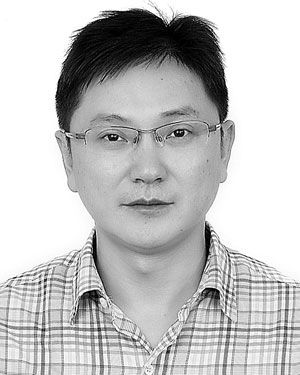}}]{WeiWei Xu}
is a researcher with the State Key Lab of CAD \& CG, College of Computer Science, Zhejiang University, awardee of NSFC Excellent Young Scholars Program in 2013. His main research interests include the digital geometry processing, physical simulation, computer vision, and virtual reality. He has published around 70 papers on international graphics journals and conferences, including 16 papers on ACM TOG. He is a member of the IEEE.
\end{IEEEbiography}

\end{document}